\documentclass{ieeeaccess}
\usepackage{cite}
\usepackage{amsmath,amssymb,amsfonts}
\usepackage{algorithmic}
\usepackage{graphicx}
\usepackage{textcomp}
\usepackage{tabularx}
\usepackage{array}
\usepackage{ragged2e}
\newcolumntype{P}[1]{>{\RaggedRight\arraybackslash\hspace{0pt}}p{#1}}

\def\BibTeX{{\rm B\kern-.05em{\sc i\kern-.025em b}\kern-.08em
    T\kern-.1667em\lower.7ex\hbox{E}\kern-.125emX}}
\begin{document}
\history{Date of publication xxxx 00, 0000, date of current version xxxx 00, 0000.}
\doi{10.1109/ACCESS.2023.0322000}

\title{Anomaly Detection in Industrial Machinery using IoT Devices and Machine Learning: a Systematic Mapping}
\author{\uppercase{S\'ergio F. Chevtchenko}\authorrefmark{1}, 
\uppercase{Elisson da Silva Rocha}\authorrefmark{1}, 
\uppercase{Monalisa Cristina Moura Dos Santos}\authorrefmark{1},
\uppercase{Ricardo Lins Mota}\authorrefmark{1},
\uppercase{Diego Moura Vieira}\authorrefmark{1},
\uppercase{Ermeson Carneiro de Andrade}\authorrefmark{2}
\uppercase{Danilo Ricardo Barbosa de Ara\'ujo}\authorrefmark{2}
}

\address[1]{SENAI Institute of Innovation for Information and Communication Technologies (ISI-TICs), Recife, Brazil)}
\address[2]{Department of Computing at the Federal Rural University of Pernambuco (UFRPE), Recife, Brazil}

\tfootnote{This work was supported by the Brazilian Agency for Industrial Development - ABDI.}

\markboth
{Chevtchenko \headeretal: Anomaly Detection in Industrial Machinery using IoT Devices and Machine Learning: a Systematic Mapping}
{Chevtchenko \headeretal: Anomaly Detection in Industrial Machinery using IoT Devices and Machine Learning: a Systematic Mapping}

\corresp{Corresponding author: S\'ergio F. Chevtchenko (e-mail: sergio.chevtchenko@sistemafiepe.org.br).}

\begin{abstract}
Anomaly detection  is critical in the smart industry for preventing equipment failure, reducing downtime, and improving safety. 
Internet of Things (IoT) has enabled the collection of large volumes of data from industrial machinery, providing a rich source of information for Anomaly Detection (AD). However, the volume and complexity of data generated by the Internet of Things ecosystems make it difficult for humans to detect anomalies manually. Machine learning (ML) algorithms can automate anomaly detection in industrial machinery by analyzing generated data. Besides, each technique has specific strengths and weaknesses based on the data nature and its corresponding systems. 
\textcolor{black}{However, a large portion of the existing systematic mapping studies on AD primarily focus  on addressing network and cybersecurity-related problems, with limited attention given to the industrial sector. Additionally, the related literature do not cover the challenges involved in using ML for AD in industrial machinery within the context of the IoT ecosystems.} 
Therefore, this paper presents a systematic mapping study on AD for industrial machinery using IoT devices and ML algorithms to address this gap. \textcolor{black}{Our primary objective is to investigate the use of ML models for anomaly detection within an industrial setting, particularly within IoT ecosystems.} The study comprehensively evaluates 84 relevant studies spanning from 2016 to 2023, providing an extensive review of AD research. Our findings identify the most commonly used  algorithms, preprocessing techniques, and sensor types. Additionally, this review  identifies application areas and points to future challenges and research opportunities. 
\end{abstract}


\titlepgskip=-21pt

\maketitle

\section{Introduction}\label{Introduction}

In recent years, many tasks previously performed by humans have been automated through smart technologies. In the Smart Industry or Industry 4.0, for example, these technologies have made it possible to monitor production lines and detect potential problems before they become serious, leading to fewer delays and increasing productivity~\cite{lee2014service,kaupp2021context}. These industries are constantly present in our daily lives and disruptions can negatively affect productivity and cause economical losses.
Generally, these disruptions come from various types of unexpected and non-standard behaviors that can occur in these environments, known as anomalies~\cite{chandola2009anomaly}. In industrial machinery, anomalies can occur due to equipment malfunctions, environmental conditions, and changes in operating conditions. Therefore, detecting these anomalies as early as possible is essential to prevent equipment failure, reduce downtime, and minimize repair costs~\cite{chandola2009anomaly}.

IoT ecosystems have enabled the collection of large volumes of data from industrial machinery, providing a rich source of information for anomaly detection~\cite{chatterjee2022iot}. 
These ecosystems are composed of sensors and monitoring systems that can collect data such as temperature, pressure, vibration, and power consumption. This data enables engineers and specialists to monitor the health of industrial machinery in real-time and take immediate action to diagnose and address the underlying problem before it causes a significant disruption to the production process. 
However, the volume and complexity of data generated by IoT ecosystems make it difficult to detect anomalies manually, increasing the number of corrective and preventive maintenance. ML algorithms can help automate the process of AD by analyzing the data generated by these environments~\cite{huch2018machine}. 
Despite the fact that there are several ML techniques used for AD, each technique has its strengths and weaknesses and can be used depending on the nature of the data and the specifics of the industrial contexts. 

Most of the current systematic mapping (SM) study on AD is in Network Anomaly Detection (NAD), considering the cybersecurity context~\cite{ahmed2016survey,eltanbouly2020machine,wang2021machine}. They are widely used for fraud detection, network security breaches, and environmental monitoring. Very few of them have focused on AD in the industry~\cite{dalzochio2020machine,zonta2020predictive}. In~\cite{dalzochio2020machine}, the authors applied a systematic literature review to identify frameworks, architectures, and tools in the area of predictive maintenance. Similarly, Zonta \textit{et al.}~\cite{zonta2020predictive} present a survey, discussing the current challenges and limitations in predictive maintenance and proposing a new taxonomy to classify this research area considering Industry 4.0. 
However, to the best of our knowledge, no related work presents and discusses the challenges in using ML for the detection of anomalies in industrial machines using IoT. 

Considering these issues and challenges described above, and the growing acknowledgment of the necessity for AD solutions, a systematic mapping study on the AD topic is highly necessary. \textcolor{black}{A mapping study offers appealing advantages, when compared to other review methods, such as a Systematic Literature Review (SLR), or a traditional narrative review. First, an SLR generally provides a deep dive into a particular question, often evaluating the quality of the studies in detail, while a SM offers a broader view, helping to understand general trends and challenges in a new field~\cite{napoleao2017practical}. Second, while traditional reviews are insightful, they may lack the rigorous methodological approach of a SM or an SLR, leading to potential biases in the covered literature.}
Therefore, this paper presents a SM of AD for industrial machinery using IoT devices and ML algorithms conducted between October 2022 and January 2023. In this study, 84 papers dating from 2016 to 2023 are evaluated. 
This SM aims to provide a comprehensive review of AD research, including an analysis of current methodologies, a synthesis of evidence, an identification of applications, a discussion of research issues, and an identification of future challenges.

This mapping study provides insights into key aspects of AD research in the context of industrial machinery, addressing the following questions: (i) What type of machinery is most commonly monitored and why?  (ii) What are the types of sensors and variables employed for detecting anomalies? (iii)  What are the types of machine learning techniques used for anomaly detection in industrial machinery? (iv) How is the anomaly detection method computed and evaluated? 
In this way, this work can help to direct future efforts regarding AD solutions, as well as promote a discussion about AD studies, their implications, and challenges for the future.

The remainder of the paper is organized as follows.
Section~\ref{sec_related_work} presents essential background concepts and existing reviews regarding the AD field. 
Section~\ref{ssec_study_process} introduces the protocol used to conduct the research.
%Section~\ref{sec_overview} presents an overview of AD solutions for industrial machinery using IoT devices and ML algorithms.
Section~\ref{sec_results} presents the findings for the established research questions.
Section~\ref{sec_OpenChallenges} presents the open challenges. 
Section~\ref{sec_limitations} presents the limitation of this mapping study.  
Finally, Section~\ref{sec_final} presents  the conclusions, the limitations, and the future work for this study.

\section{Background and Related Work}
\label{sec_related_work}
\textcolor{black}{In this section, we introduce key background concepts relevant to the industrial context and provide an overview of related literature while critically comparing it with our work.}

\subsection{Background}

\subsubsection{Internet of Things} 
\textcolor{black}{IoT is commonly seen as a network for data exchange among machines, primarily driven by consumer demands, with a focus on machine-to-user interactions and client-server dynamics~\cite{bandyopadhyay2011internet}. 
On the other hand, Industrial IoT (IIoT), a crucial concept in Industry 4.0, centers on connecting industrial assets, such as machinery and control systems, to business processes and information systems. The incorporation of IoT devices enhances productivity by enabling connections and data exchange within production systems~\cite{lu2017industry}. 
Furthermore, IIoT allows cost-effective upgrades of existing industrial infrastructure by adding sensors to older equipment~\cite{cunha2021upgrading}. }

\textcolor{black}{When it comes to data volume, IIoT frequently involves extensive data exchanges, facilitating machine learning-based analysis and maintenance enhancements. For instance, Ayvaz and Alpay~\cite{ayvaz2021predictive} exemplify this approach, utilizing a diverse array of built-in IoT sensors that transmit a substantial volume of data via the MQTT protocol for ML-based anomaly detection. However, it is worth noting that IIoT applications often face constraints in terms of power and computing resources~\cite{sisinni2018industrial}. In our manuscript, we provide an overview and rationale regarding the types of sensors and machinery commonly utilized in the industrial domain. 
%Additionally, we point that the current literature largely lacks experimental data involving real-time performance and computational cost of the anomaly detection approaches.
}
\subsubsection{Machine Learning and Anomaly Detection}
\textcolor{black}{
Machine learning is a tool used to enable computers to perform complex tasks like prediction, diagnosis, planning, and recognition. 
The success of ML relies on the availability of high-quality data and large datasets for algorithm training, both of which have become more readily accessible in the industrial context following the advent of IIoT~\cite{angelopoulos2019tackling}. 
ML techniques can be classified in supervised, unsupervised, and reinforcement learning~\cite{wuest2016machine}. In our study, we observe the application of both supervised and unsupervised approaches to anomaly detection. However, unsupervised algorithms present an attractive advantage by not necessitating labeled data for training. This is particularly advantageous for detecting faults in machinery, given that fault data tend to be infrequent and sparse.
}

\begin{figure}[!htbp]
    \centering
    \includegraphics[width=\columnwidth]{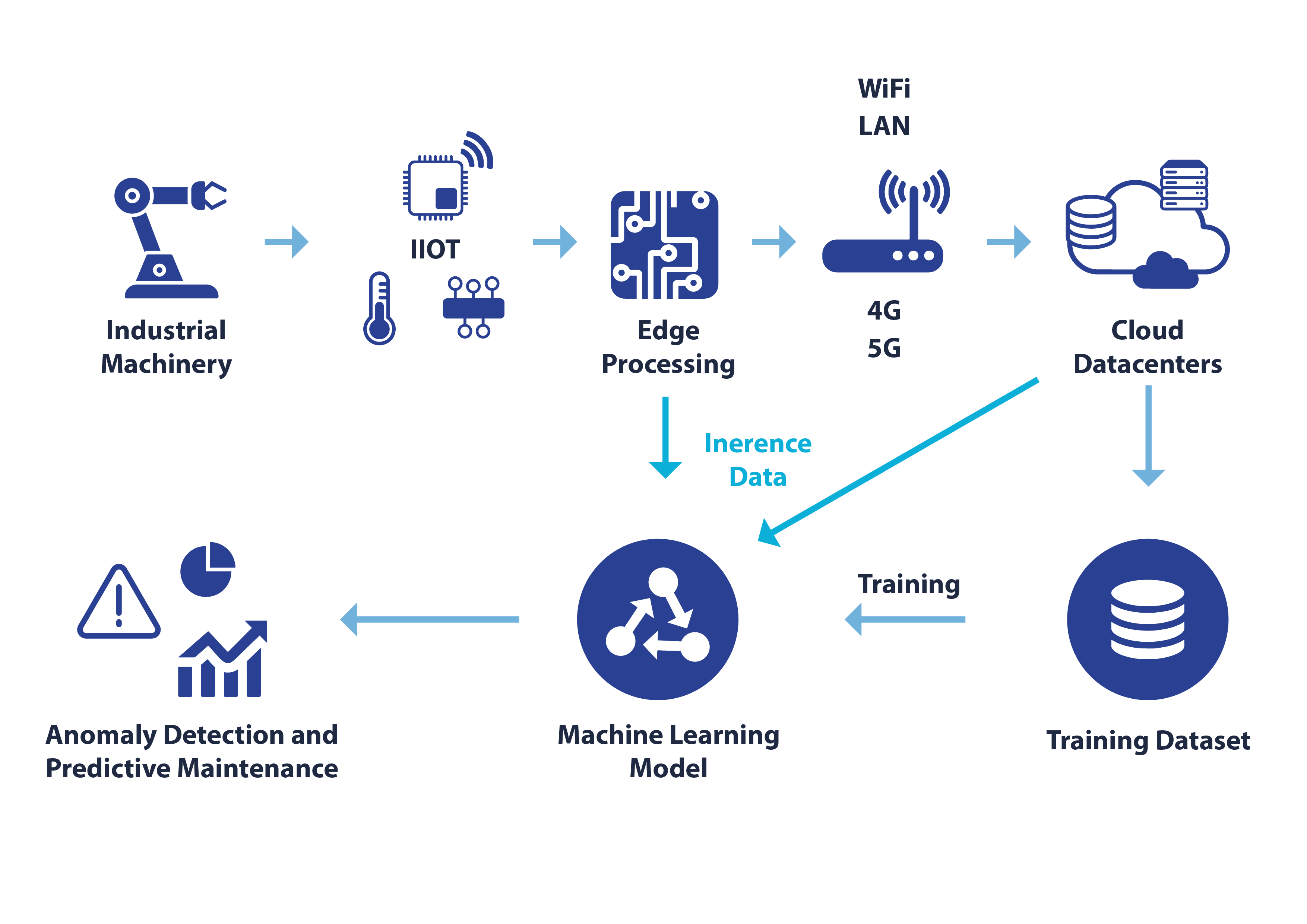}
    \caption{\textcolor{black}{An overview of the application of IIoT and ML for anomaly detection and predictive maintenance in the industry.}}
    \label{fig_overview_AD_ML}
\end{figure}

\textcolor{black}{Figure \ref{fig_overview_AD_ML} provides an overview of a practical application of IIoT and ML within an industrial context for anomaly detection and predictive maintenance.
In this application,  IIoT devices can be considered as various sensors with integrated local connectivity. Optionally, some of the generated data can be processed on devices close to the IIoT sensors, known as edge processing. This can reduce the amount of data sent to the cloud, increase the detection speed and reduce the computational cost of anomaly detection. However, our study indicates that many advanced anomaly detection algorithms require substantial computational power and, therefore, necessitate further optimization to effectively implement edge processing. 
The gathered data is then typically stored on a cloud data center and used to train AD algorithms. Many ML algorithms require extensive training on large datasets. Following the training process, the AD models can be deployed either at the edge or in the cloud for anomaly detection. Finally, real-time data can be streamed to the cloud for analysis or processed directly on edge devices to detect anomalies.
}

\subsection{Related Work}

In recent years, there has been considerable interest in AD for industrial machinery using IoT devices and ML. However, the majority of studies have primarily focused on network anomaly detection. To position our paper and emphasize its contributions, we start by presenting a concise introduction to mapping studies that concentrate on network anomaly detection. 
Subsequently, we provide a summary of previous research on surveys conducted on AD and predictive maintenance in the industry, employing ML techniques. Finally, we present a comparative analysis of our work in relation to these studies, considering factors such as scope and evaluation methods.

There is a significant body of literature comprising systematic mapping studies that specifically concentrate on NAD \cite{kumar2022intrusion, gupta2022cyber-multi}. Ahmed \textit{et al.} \cite{ahmed2016survey} explore various NAD techniques and elaborate on the use of different categories of detection methods as solutions for this problem. The authors also discuss the limitations of using publicly available intrusion detection datasets and provide a comparison of the effectiveness of presented AD techniques based on specific criteria.
In a more recent work, Eltanbouly \textit{et al.} \cite{eltanbouly2020machine} examine ML algorithms applied to NAD and analyze the performance of the surveyed papers. The study provides an overview of their main positive and negative characteristics, along with numerical analyses of algorithms learned on the same datasets.
Wang \textit{et al.} \cite{wang2021machine} discuss different types of ML approaches in the context of AD, comparing their merits within the scope of NAD. The authors also address the key challenges faced in detecting various anomalies in networks with varying complexities. Additionally, they present use cases for different types of networks, considering IoT networks, though with limited scope.

Very few mapping studies have focused on AD and predictive maintenance in the industry. Gaddam \textit{et al.} \cite{gaddam2019anomaly} investigate AD techniques to identify sensor faults and outliers in the IoT context, without specific emphasis on industrial equipment. 
Their survey comprehensively describes the primary sources of sensor outliers within the IoT environment and discusses detection models that are suitable for IoT systems. The authors further provide a comparative analysis of these techniques, highlighting their respective strengths and weaknesses.
With a focus on Industry 4.0, Kamat and Sugandhi \cite{kamat2020anomaly} provide a discussion about the main challenges  associated with traditional equipment maintenance strategies in the manufacturing industry. 
As a way to address this problem, the study suggests the adoption of AD in predictive maintenance as a more suitable approach for handling the specific data characteristics present in this environment. 
Additionally, the paper briefly explores alternative approaches to AD found in the literature and discusses publicly available datasets relevant to the manufacturing scenario.
 
Dalzochio \textit{et al.} \cite{dalzochio2020machine} present important considerations regarding the  application of predictive maintenance, such as the abundance of data, the criticality of the equipment, the need for redundancy and the availability of failure-related data for modeling purposes. The researchers address several key research questions, including the challenges associated with applying ML to predictive maintenance, commonly employed ML techniques in this context, and the utilization of ontologies in predictive maintenance scenarios. In terms of ML techniques, the analyzed works were grouped into three main categories: (i) based on artificial neural networks (ANN); (ii) based on deep learning (DL); and (iii) based on other ML approaches, such as k-nearest neighbors (kNN), support vector machines (SVM), and random forest (RF). The authors suggest that different ML models are better suited for AD/fault classification and prognostics. While classification or clustering ML models can address AD and fault classification, regression models such as autoregressive integrated moving averages (ARIMA) are better suited for prognostics tasks. Furthermore,  the paper provides a concise exploration of the use of ontologies in predictive maintenance, aiming to facilitate decision-making processes.

Zonta \textit{et al.} \cite{zonta2020predictive} present a systematic literature review that explores initiatives related to predictive maintenance in the context of Industry 4.0. The authors analyze 47 articles and, as the main contribution, propose a taxonomy for predictive maintenance in the context of Industry 4.0. 
Schwendemann \textit{et al.} \cite{schwendemann2021survey} conduct a survey that specifically explores ML techniques for predictive maintenance of bearings in grinding machines. The selected works encompass a range of approaches, including ANNs, Hidden Markov Models (HMMs), and SVMs. 

Kang \textit{et al.} \cite{kang2020machine} conduct a systematic literature review that specifically examines the applications of ML in production lines and their components. Their analysis included 39 primary studies that predominantly concentrated on quality control within the production lines. The most frequently utilized dependent variables in the ML methods were identified as Fail/Pass indicators, physical properties of materials, and object dimensions.

With a focus on the railway industry, Davari \textit{et al.} \cite{davari2021survey} present a survey on data-driven predictive maintenance. Within this context, they emphasize real-time AD from time-series data and the need for automatic reasoning capabilities to explain causality as major challenges faced by the industry. 
Nor \textit{et al.} \cite{nor2021overview} present an analysis with a specific focus on explainable artificial intelligence (XAI) applied to prognostics and health management (PHM) of industrial assets. The authors adopt the preferred reporting items for systematic reviews and meta-analyses (PRISMA) methodology in their study. They analyzed a total of 35 selected works, with 13\% of them incorporating an AD approach. 
It is worth highlighting that the analyzed studies primarily feature case studies based on real industrial data. This demonstrates the practical application of AI models in industrial settings and contributes to an increasing level of confidence in adopting such models within the industry.

\textcolor{black}{Our work distinguishes itself from the related literature above for the following reasons. Firstly, AD-based predictive maintenance is recognized as a promising area in various broader literature reviews. Thus, our work offers a state-of-the-art review with a focused approach on the use of ML techniques for detecting anomalies in industrial machinery. To the best of our knowledge, this is the first mapping study to concentrate specifically on anomaly detection in the industrial context.
Secondly, unlike previous studies, we do not limit our analysis to specific industrial machinery, aiming to provide a more comprehensive perspective on existing industrial practices. As a result, our work seeks to offer an up-to-date review of the state-of-the-art, with a specific emphasis on the application of machine learning methods for industrial anomaly detection.
Finally, the present work incorporates the most recent findings from the literature, ensuring that our review is current and reflective of the latest advancements in the field. 
The points mentioned above are further elaborated in Table \ref{tab:comparison_related}, where our work is compared to the related literature. 
The ``Type'' column denotes the methodology of the referenced research. ``Survey'' generally implies a more qualitative review. ``SLR'' indicates a more rigorous and structured approach to reviewing existing literature, whereas ``SM'' offers are . The ``Research Period'' column indicates the time span the study covered. The ``IoT'' column indicates the extent to which the study considered the IoT. ``Yes'' indicates the study contains significant focus on IoT considerations, ``Partial'' implies limited inclusion of IoT in a broader focus of the research, and ``No'' means that the study did not focus on IoT to any meaningfull extent. Finally, the ``Industrial Application'' column indicates whether the study has a wider and more general application within the industry,  or a narrower focus like ``Beargins'' or ``Railway''.}

\begin{table}[!htbp]
\caption{\textcolor{black}{A comparison of related works with the present study.}}
\label{tab:comparison_related}
\begin{tabular}{ccccc}
\hline
Work  & Type & \begin{tabular}{@{}c@{}}Research \\ period\end{tabular} & IoT & \begin{tabular}{@{}c@{}}Industrial \\ application\end{tabular}  \\ \hline
\cite{gaddam2019anomaly} & Survey & up to 2019 & Yes & No \\
\cite{kamat2020anomaly} & Survey & up to 2019 & Partial & Broad \\
\cite{dalzochio2020machine} & SLR & 2015-2020 & No & Broad \\
\cite{zonta2020predictive} & SLR & 2008–2018 & Yes & Broad \\
\cite{schwendemann2021survey} & Survey & up to 2020 & Yes & Bearings \\
\cite{kang2020machine} & SLR & 2014–2019 & Partial & \begin{tabular}{@{}c@{}}Production \\ lines\end{tabular} \\
\cite{davari2021survey} & Survey & up to 2020 & Partial & Railway \\
\cite{nor2021overview} & Survey & up to 2020 & Partial & Broad \\
Our work & SM & 2012–2022 & Yes & Broad \\
\hline
\end{tabular}
\end{table}

\section{Systematic Mapping Study Process}
\label{ssec_study_process}

\begin{figure}[!htbp]
    \centering
    \includegraphics[width=\columnwidth]{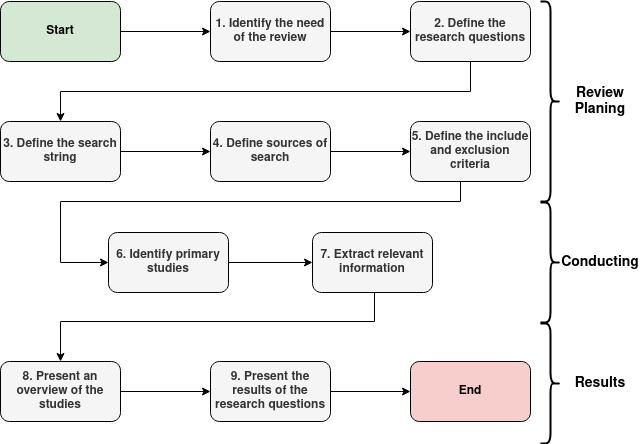}
    \caption{Steps of the adopted protocol.}
    \label{fig:methodology}
\end{figure}

Our systematic mapping was guided by the methodology proposed by Petersen \textit{et al.}~\cite{petersen2008systematic}, which we employed to identify articles related to the use of ML techniques for detecting anomalies in industrial machinery with the aid of IoT devices, as illustrated in Figure \ref{fig:methodology}.

\emph{Step 1. } Given the constantly growing volume of data generated by industrial machinery, the use of ML models has become increasingly significant in detecting anomalies and enabling preventive maintenance, as opposed to corrective maintenance. 
The primary objective of this SM is to investigate the application of ML models for predicting anomalies in industrial machinery in the context of IoT ecosystems. Specifically, this study aims to identify the most commonly adopted machinery, explore the types of data and sensors utilized for AD, identify the computation and evaluation methods used for AD, and analyze the challenges and opportunities associated with implementing ML-based AD systems in industrial settings. The insights gained from this research will be critical in assessing the current state of this field and providing guidance for future research endeavors in this domain.

\emph{Step 2. }Our SM research began by identifying key Research Questions (RQs) aimed at exploring the current state of AD in industrial machinery using ML algorithms and sensors. The following RQs were adopted for this SM:

\begin{itemize}
    \item RQ1: What type of machinery is most commonly monitored and why? 
    \item RQ2: What are the types of sensors and variables employed for detecting anomalies?
    \item RQ3: What are the types of machine learning techniques used for anomaly detection in industrial machinery?
    \item RQ4: How is the anomaly detection method computed and evaluated? 
\end{itemize}

\emph{Step 3. }To compile primary research, we employed automatic database searches utilizing a search string. The following procedures were undertaken to elaborate the search phrase for the automatic search \cite{mendonca2019disaster, kitchenham2007cross}: (i) extract phrases from the study questions; (ii) find alternate spellings and synonyms; (iii) validate the keywords; and IV. combine search strings using Boolean operators (OR, AND). The search string was (("Anomaly Detection" OR "Anomalous Behavior") AND ("IoT" OR "Internet of Things" OR "Sensors") AND ("Machine Learning" OR "Deep Learning" OR "Artificial Intelligence")).

\emph{Step 4. }The following  digital libraries were considered as the primary sources for our research:
IEEE Xplore\footnote{IEEExplore.ieee.org/Xplore/home.jsp}, ACM Digital Library\footnote{https://dl.acm.org}, Science Direct\footnote{https://www.sciencedirect.com/} and Web of Science\footnote{https://www.webofscience.com}.

\emph{Step 5. }As numerous papers unrelated to our research topics may be found, we have established specific inclusion and exclusion criteria. These criteria are intended to narrow down our search and ensure that the identified literature is relevant to our assessment research. 
For the inclusion criteria, we focused on papers published within the last ten years that explicitly address the AD using ML and IoT sensors in their abstracts. We excluded, on the other hand, duplicate, unavailable, or foreign-language articles, as well as editorials, posters, tutorials, and secondary or tertiary articles.

\emph{Step 6. }After applying the inclusion and exclusion criteria and utilizing the search string in the digital libraries, we successfully located the primary studies. 

\emph{Step 7. }We extracted pertinent information from the primary studies by thoroughly reading the entire paper and answering the RQs.

\emph{Step 8. }To classify and organize the articles in accordance with our research questions stated in Step 2, an overview of all articles is provided in this stage (see Subsection \ref{sec_overview}).

\emph{Step 9. }Finally, in Section \ref{sec_results}, we present the responses we found to the RQs posed in Step 2, which details the current state of the literature regarding the utilization of ML and sensors to identify anomalies in industrial machinery.

\section{Results and Discussion}
\label{sec_results}

 To provide a comprehensive understanding of AD techniques in industrial machinery, we begin with an overview of the primary studies conducted in this field. We then present specific issues related to AD, such as the types of machinery commonly monitored, the sensors and variables employed for detecting anomalies, the types of ML techniques used for AD, and how the AD methods are computed and evaluated. This approach provides a solid understanding of the techniques and challenges involved in detecting anomalies in industrial machinery.

\subsection{Overview of the Primary Studies}
\label{sec_overview}

The number of studies identified by the search string in the search sources, as well as the number of studies following the application of the inclusion and exclusion criteria, are presented in Table~\ref{tab:search-results}. Initially, 8966 articles were retrieved from the 4 databases. After removing duplicates, the number decreased to 8037 articles. Through the application of the inclusion and exclusion criteria, 84 articles were selected for thorough reading and to address the research questions. This represents only 0.936\% of all articles found during the search, emphasizing the importance of applying rigorous selection criteria to identify relevant studies.

\begin{table}[!htbp]
\caption{Search results obtained before and after applying the inclusion and exclusion criteria.}
\label{tab:search-results}
\begin{tabular}{ccc}
\hline
Database       & Original Search & \begin{tabular}{@{}c@{}}\emph{After Primary} \\ \emph{Studies Identification}\end{tabular}  \\ \hline
ACM DL         & 1713            & 6                                    \\
IEEE Xplore    & 979             & 11                                   \\
Web of Science & 1249            & 27                                   \\
Science Direct & 5025            & 40                                   \\ \hline
\end{tabular}
\end{table}

Among the databases searched Science Direct yielded the highest number of selected works, with 40 articles. This was followed by Web of Science with 27, IEEE Xplore with 11, and finally ACM DL with only 6 articles. In terms of the proportion between the number of selected articles and the original search, Web of Science achieved the best result with approximately 2.16\% (27/1249) of the selected base, followed by IEEE Xplore with 1.12\% (11/979).

\begin{figure}[!htbp]
    \centering
    \includegraphics[width=0.9\columnwidth]{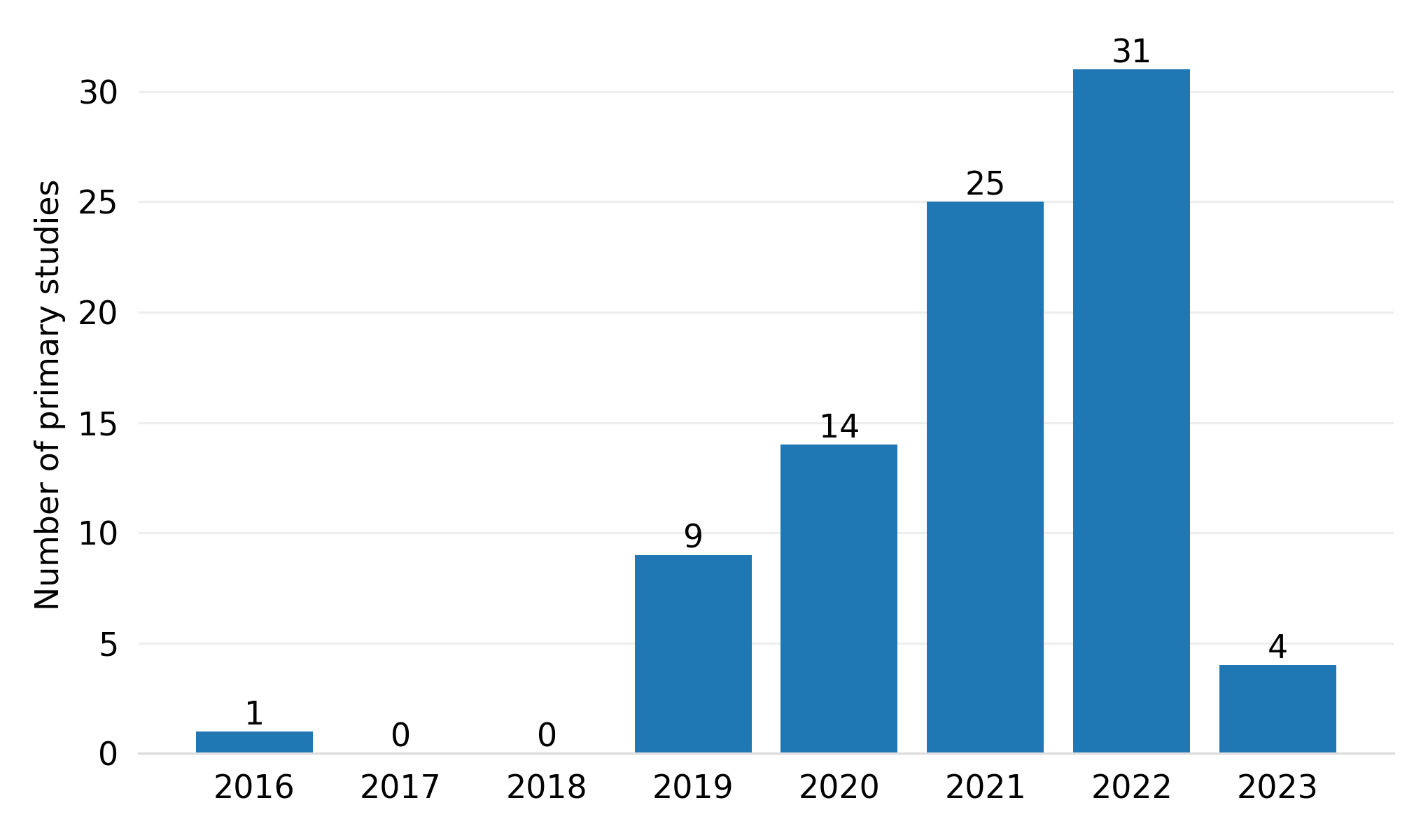}
    \caption{Number of primary studies by publication year.}
    \label{fig:primary-studies-year}
\end{figure}

Based on the selected articles, Figure~\ref{fig:primary-studies-year} presents the number of primary studies in relation to the year of publication. Although our search aimed to include articles from 2012 onwards, the first selected study was from 2016. Notably, no articles were selected for 2017 and 2018. However, we observed a growing trend in the literature regarding the detection of anomalies in industrial machinery using sensors and ML from 2019 onwards. In 2019, 9 studies were selected, increasing to 21 in 2020, 28 in 2021, and reaching 39 in 2022. It is worth mentioning that even though our search was conducted up until November 2022, some studies have already been published in 2023, indicating that research in this area is ongoing.

\begin{figure}[!htbp]
    \centering
    \includegraphics[width=\columnwidth]{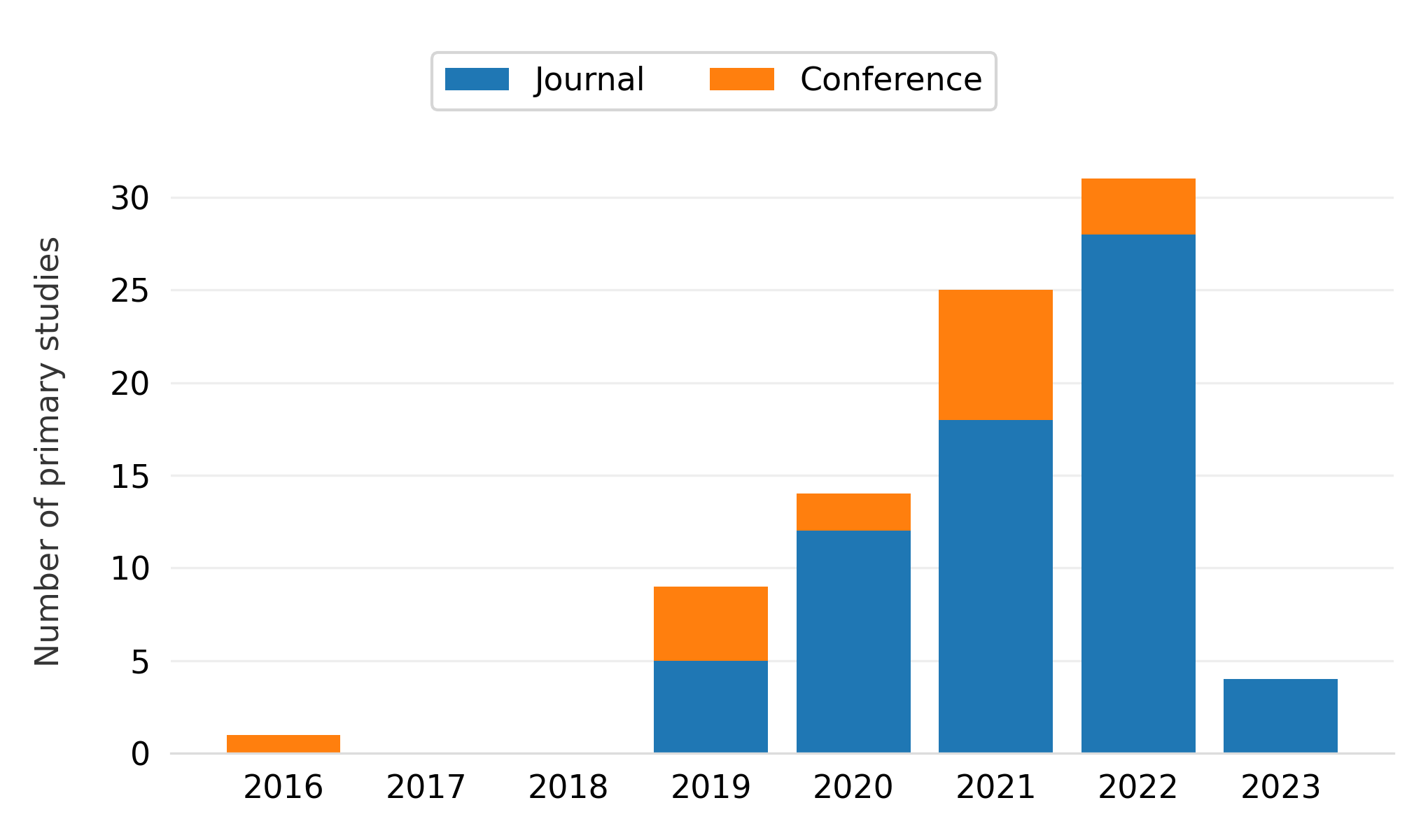}
    \caption{Number of primary studies by year and publishing type.}
    \label{fig:primary-studies-year-pub-venue-type}
\end{figure}

Based on the type of publication, journals represent the majority, comprising 80\% of primary studies, while conferences account for the remaining 20\%. We can also examine the trends in publication venues over the years  (see Figure~\ref{fig:primary-studies-year-pub-venue-type}). The first publication, in 2016, was presented at a conference. By 2019, the proportion of journal publications had increased significantly, with 55\% of the publications (5 papers) being in journals, and 45\% (4 papers) in conferences. In the subsequent years, the number of journal publications continued to rise, with 12 papers in 2020, 18 in 2021, and 28 in 2022. This trend could be attributed to multiple factors, including the pandemic, which prompted researchers to shift their focus to publishing in journals rather than presenting at conferences. Additionally, the natural progression and development of research studies over time may have contributed to the increase in journal publications.

Several journals have published a significant number of works on the topic of this MS, with SENSORS being the most productive with 9 publications, followed by Procedia CIRP with 5 publications. Expert Systems with Applications, IEEE Access, Journal of Manufacturing Systems, and Mechanical Systems and Signal Processing also had a substantial output, each with 4 publications. All of these journals have a notably high impact factor. It is worth mentioning that none of the publications were presented at the same conference,  highlighting the diversity of the research presented and the scope of the academic community.

\subsection{What type of machinery is most commonly monitored and why?}
\label{subsec_RQ1}

\begin{figure}[!htbp]
    \centering
    \includegraphics[width=\columnwidth]{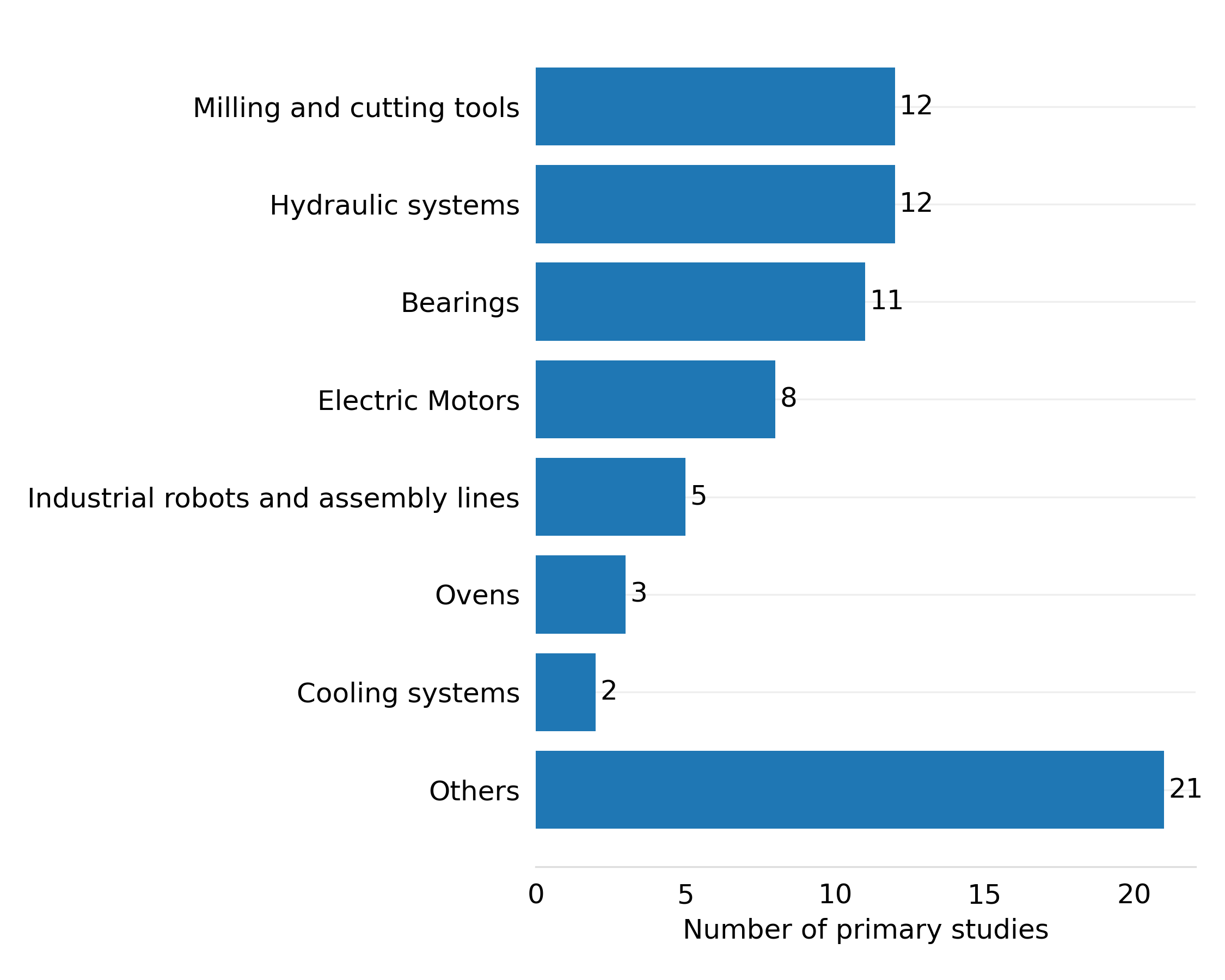}
    \caption{Number of primary studies by the type of industrial machinery.}
    \label{fig:machinery}
\end{figure}

To answer the above research question, we listed the articles that clearly indicate the type of machinery under evaluation. From this selection, we created 11 groups based on categorical proximity and their use in the articles (e.g., articles that mentioned pumps also mentioned valves and hydraulic systems. In these cases, we considered hydraulic systems). The number of works gathered in each group is shown in Figure~\ref{fig:machinery}. The machinery that did not fit into any existing group was not used more than once, were placed in a group called ``others''.
For this analysis, studies in which the monitored machinery could not be identified were excluded.
Based on the number of occurrences in each group, we ranked the groups to observe the types of machinery most monitored in the primary studies.

As shown in Figure~\ref{fig:machinery}, the three primary types of machinery used for anomaly detection are milling and cutting tools, hydraulic systems, and bearings. Several studies conducted across these equipment groups have implemented predictive maintenance techniques with the goal of achieving various benefits. These benefits include reducing the amount of lost production, improving the quality of products, minimizing the costs associated with unexpected equipment downtime, and increasing overall production efficiency. Furthermore, the implementation of predictive maintenance can also lead to an improvement in the lifespan of the machinery.
\cite{jakubowski2021anomaly, hong2022intelligent, assafo2021topsis, zabinski2021fpga}%\cite{jakubowski2021anomaly, hong2022intelligent, assafo2021topsis, zabinski2021fpga, ryu2022quantile, guo2022unsupervised, zhai2021enabling, watanabe2020anomaly, ying2021hybrid, yan2023hybrid, zufle2022workflow, feng2021time, abbasi2021outliernets, boons2021edge, di2022anomalous, tagawa2021acoustic, li2020lifelong, siegel2020industrial, de2022long, mostafavi2021novel, kim2022anomaly, keleko2023health, lindemann2019anomaly, bose2019adepos, pittino2020automatic, brito2022explainable, luo2022multi, liu2021online, conradi2021anomaly, song2019anomaly}.

In the category of milling and cutting tools, numerous studies concentrate on assessing the effects of wear and tear on their surfaces. These studies are particularly relevant for machinery such as Computer Numerical Control (CNC) routers, milling machines, lathes, and hot rolling mills, which frequently handle heavy-duty tasks involving cutting metal, ceramics, and other hard materials. The nature of these tasks makes these tools susceptible to significant wear, which is a primary reason for continuous monitoring~\cite{jakubowski2021anomaly, hong2022intelligent, assafo2021topsis, zabinski2021fpga, ryu2022quantile, guo2022unsupervised, zhai2021enabling, watanabe2020anomaly}.
As these milling and cutting tools wear out due to intensive usage, they can directly impact the quality of manufactured products and even cause damage to the machinery itself. In some cases, routine maintenance involves replacing these worn-out parts, even when it might still be possible to continue using them~\cite{jakubowski2021anomaly}.

Hydraulic systems consist of several components, such as accumulators, coolers, valves, pumps, compressors, and more. Depending on the application, these systems can operate either as standalone units or as part of larger systems. For example, in aviation, hydraulic systems play a crucial role in controlling mechanical parts of aircraft, while in other applications, they may be integrated into cooling systems~\cite{kim2022anomaly}. It is important to note that some hydraulic systems operate under high pressure and deal with toxic or flammable fluids, emphasizing the criticality of proper maintenance for safety. Traditionally, manual monitoring of these hydraulic systems has been employed, relying on the extensive knowledge and judgment of operators. However, this approach can be subjective and may lead to inefficiencies. Therefore, researchers advocate for the adoption of automatic monitoring techniques using IA models to detect anomalies in hydraulic systems~\cite{abbasi2021outliernets, boons2021edge, di2022anomalous, tagawa2021acoustic}.

Bearings are the third most monitored component in industrial machinery. Due to their critical role in supporting rotating parts and reducing friction, monitoring the condition of bearings is of utmost importance to ensure the smooth operation and longevity of various machines and equipment~\cite{juodelyte2022predicting,song2019anomaly}. For instance, a recurrent issue arises when wind turbines operate at low speeds, subjecting the bearings to significant stress due to the considerable weight of the turbine's components~\cite{konig2021machine}. Predicting abnormal behavior in bearing wear is essential in such scenarios.
Furthermore, in this mapping study, several other machinery were identified, including electric motors and industrial robots. However, milling and cutting tools, hydraulic systems, and bearings were found to be prevalent.

Given their extensive use in almost all industrial environments, it is imperative for these machines to operate smoothly and consistently even under demanding conditions for extended periods.  Consequently, predicting anomalous behavior in industrial machinery is essential to avoid negative impacts on productivity, as it allows for the early detection of potential faults and the implementation of preventive maintenance measures. It is worth highlighting that many of the analyzed studies do not focus on evaluating specific machinery but rather on exploring new ML and DL approaches using datasets containing data related to various machinery types. Nonetheless, they contribute to the overall understanding of AD techniques and their applicability in the context of industrial machinery.

%\cite{ryu2022quantile, guo2022unsupervised, zhai2021enabling, watanabe2020anomaly, ying2021hybrid, yan2023hybrid, zufle2022workflow, feng2021time, abbasi2021outliernets, boons2021edge, di2022anomalous, tagawa2021acoustic, li2020lifelong, siegel2020industrial, de2022long, mostafavi2021novel, kim2022anomaly, keleko2023health, lindemann2019anomaly, jiang2022multiscale, shi2021deep, juodelyte2022predicting, konig2021machine, bose2019adepos, pittino2020automatic, brito2022explainable, luo2022multi, liu2021online}.

\subsection{What are the types of sensors and variables employed for detecting anomalies?}
\label{subsec_RQ2}

The majority of the primary studies focused on AD modeling and, as a result, lacked detailed technical information on the types of sensors used for machinery monitoring. This focus on AD over the whole system monitoring pipeline, which includes data acquisition, processing, and information delivery, limits the depth of our analysis. Moreover, many studies validated the proposed methodologies using well-established datasets, such as C-MAPSS\footnote{https://data.nasa.gov/dataset/C-MAPSS-Aircraft-Engine-Simulator-Data/xaut-bemq}~\cite{saxena2008damage} and the NASA Ames Prognostic Data Repository\footnote{https://www.nasa.gov/content/prognostics-center-of-excellence-data-set-repository}~\cite{assafo2021topsis, bose2019adepos, nguyen2023time, jakubowski2021anomaly}. While these datasets serve as benchmarks for experimental purposes, it is important to test these methodologies in real-world industrial settings for practical applications.
Despite these limitations, the majority of the studies did reveal the variables used for detecting AD. We have categorized these variables based on the type of sensor used, such as vibration, electric current, temperature, noise, pressure, among others. Table~\ref{tab_sensors_machines} lists the studies that employed each sensor type, either individually or in combination with each other.

\begin{table*}[ht]
\caption{Sensors used in selected works.}
\label{tab_sensors_machines}
\scriptsize
\renewcommand{\arraystretch}{1.5}
\begin{tabularx}{\textwidth}{P{0.4\textwidth}ccccccc}
\hline
\textbf{Article} & \textbf{Vibration} & \textbf{Electric Current} & \textbf{Temperature} & \textbf{Noise} & \textbf{Pressure} & \textbf{Others} \\ 
\hline
\cite{patra2022anomaly,mostafavi2021novel,jiang2022multiscale,yu2022fastadaptation,kilian2022vibration,brito2022explainable, jian2021initial, juodelyte2022predicting,de2022hybrid,zabinski2021fpga,bose2019adepos,li2020lifelong,pittino2020automatic,alfeo2020using} & \checkmark &  &  &  &  &  \\ 
\cite{yun2020development,ying2021hybrid,meire2019comparison,tagawa2021acoustic,abbasi2021outliernets,boons2021edge,di2022anomalous,ahn2021deep} &  &  &  & \checkmark &  &  \\ 
\cite{gruner2020evaluation,hiruta2021unsupervised,watanabe2020anomaly,givnan2022anomaly,park2021design} &  & \checkmark &  &  &  &  \\ 
\cite{panagou2022feature,yanabe2020anomaly,gonzalez2022two,velasquez2022hybrid,kim2022anomaly} & \checkmark & \checkmark & \checkmark &  & \checkmark & \checkmark \\
\cite{canizo2019multi,conradi2021anomaly,yan2023hybrid,denkena2021data} & \checkmark &  &  &  &  & \checkmark \\
\cite{yang2022interpretable,apostol2021change,lindemann2019anomaly} &  &  &  &  & \checkmark & \checkmark \\
\cite{meyer2022anomaly,kammerer2019anomaly,langone2020interpretable} &  & \checkmark & \checkmark &  & \checkmark & \checkmark \\
\cite{ayvaz2021predictive,calvo2023collaborative,hong2022intelligent} & \checkmark & \checkmark & \checkmark &  &  &  \\
\cite{abbracciavento2021anomaly,wang2016self} &  &  & \checkmark &  &  &  \\
\cite{guo2022unsupervised,nguyen2023time} & \checkmark & \checkmark &  & \checkmark &  & \checkmark \\
\cite{wielgosz2019mapping,ryu2022quantile} &  & \checkmark &  &  &  & \checkmark \\
\cite{zhai2021enabling,jakubowski2021anomaly} &  &  & \checkmark &  & \checkmark & \checkmark \\
\cite{keleko2023health,coelho2022predictive} & \checkmark &  & \checkmark &  & \checkmark & \checkmark \\
\cite{tancredi2022integration,de2022long} &  &  &  &  & \checkmark &  \\
\cite{liu2021online,antonini2022tinyml} & \checkmark &  & \checkmark &  &  &  \\
\cite{zufle2022workflow,yasaei2020iot} & \checkmark &  & \checkmark & \checkmark &  & \checkmark \\
\cite{song2019anomaly} &  & \checkmark & \checkmark &  &  & \checkmark \\
\cite{konig2021machine} &  &  & \checkmark & \checkmark &  & \checkmark \\
\cite{de2020use} & \checkmark & \checkmark & \checkmark & \checkmark &  & \checkmark \\
\cite{hendrickx2020general} & \checkmark & \checkmark &  &  &  &  \\
\cite{kim2019online} &  &  & \checkmark &  &  & \checkmark \\
\cite{calvo2021anomaly} &  & \checkmark & \checkmark &  &  &  \\
\cite{lu2021gan} &  & \checkmark &  & \checkmark &  &  \\
\cite{shi2021deep} & \checkmark & \checkmark &  &  & \checkmark & \checkmark \\
\cite{ds2022comparative} &  &  & \checkmark &  & \checkmark &  \\
\cite{luo2022multi} & \checkmark & \checkmark & \checkmark &  &  & \checkmark \\
\cite{choi2022explainable} & \checkmark & \checkmark & \checkmark &  & \checkmark &  \\
\cite{assafo2021topsis} & \checkmark & \checkmark &  & \checkmark &  &  \\
\cite{qian2021multichannel,iftikhar2020outlier,wang2021early,wu2019lstm,wang2022detecting,netzer2022process,lughofer2020line} &  &  &  &  &  & \checkmark \\
\hline
\end{tabularx}
\end{table*}

Out of the 84 examined papers, 24 relied on a single type of sensor for detecting AD. Ten of these studies exclusively used vibration sensors \cite{patra2022anomaly,mostafavi2021novel,jiang2022multiscale,yu2022fastadaptation,kilian2022vibration,brito2022explainable, jian2021initial,
juodelyte2022predicting,de2022hybrid,zabinski2021fpga,bose2019adepos,li2020lifelong,pittino2020automatic,alfeo2020using}. Seven studies depended solely on noise sensors \cite{yun2020development,ying2021hybrid,meire2019comparison,tagawa2021acoustic,abbasi2021outliernets,boons2021edge,di2022anomalous,ahn2021deep}. Five papers utilized only electric current sensors \cite{panagou2022feature,yanabe2020anomaly,gonzalez2022two,velasquez2022hybrid,kim2022anomaly}, while two used only temperature sensors \cite{abbracciavento2021anomaly,wang2016self}. Finally, two papers solely relied on pressure sensors~\cite{tancredi2022integration,de2022long}. This suggests that roughly 28.57\% of the selected studies focus on a single type of sensor. 

While there are certain cases where a single sensor may be sufficient, there are potential benefits in using multiple sensor types in the field of AD \cite{li2022correlation}. As an example, multiple sensors can be used to infer new information about the physical system under observation. Through the combined analysis of data from these varied sensors, one can extrapolate novel information not directly observed.  This technique is referred to as the ``virtual sensors'' in the literature~\cite{kabadayi2006virtual}. 
Nevertheless, many studies employed a combination of sensors, with five works opting to use the vibration, electric current, temperature and pressure sensors, along with an additional unclassified sensor type~\cite{panagou2022feature,yanabe2020anomaly,gonzalez2022two,velasquez2022hybrid,kim2022anomaly}. In three studies vibration, electric current and temperature sensors are combined~\cite{ayvaz2021predictive,calvo2023collaborative,hong2022intelligent}.  

\begin{figure}[!htbp]
    \centering
    \includegraphics[width=0.9\columnwidth]{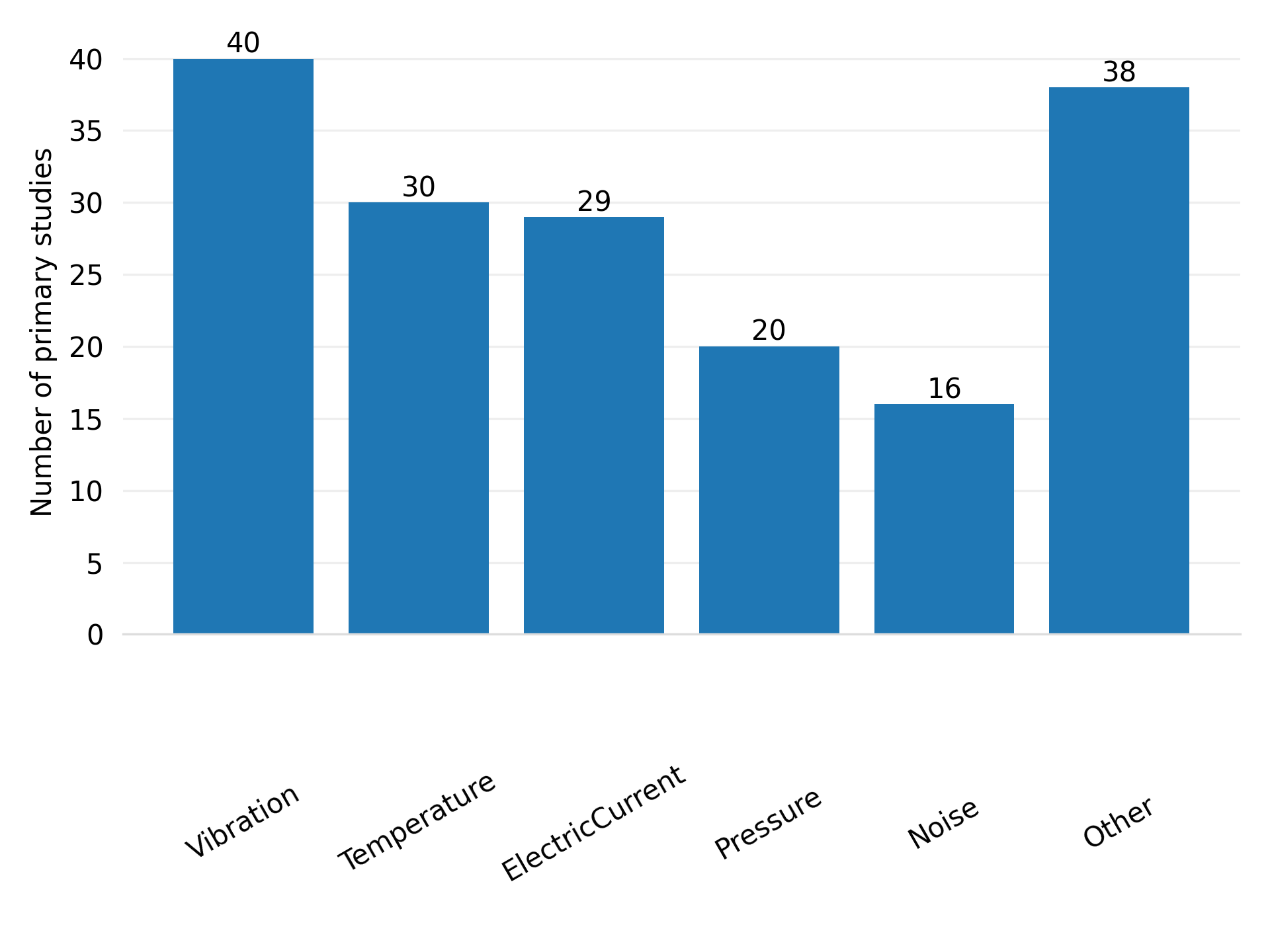}
    \caption{Number of primary studies by sensors type.}
    \label{fig:sensors-type}
\end{figure}

As depicted in Figure~\ref{fig:sensors-type}, the distribution of sensor types across the primary studies reveals a clear preference for certain sensors. The most frequently employed sensor was the vibration sensor, utilized in 40 studies. Temperature sensors were the second most common, appearing in 30 studies, closely followed by electric current sensors, which were used in 29 studies. Pressure sensors found application in 20 studies, while noise sensors were incorporated in 16 studies. A variety of other sensor types, accounting for variables such as speed, torque, RPM, humidity, viscosity, and proximity, were collectively used in 38 studies.

Despite noise sensors being the exclusive focus in seven studies as presented earlier, they were, in fact, the least frequently used when we consider the overall usage of sensor types across all studies, with just 16 occurrences. This suggests that a mix of sensor types is proffered to relying solely on noise sensors or other single sensor type.
It is worth mentioning that cameras, while traditionally used for tasks like segmentation, edge detection, object classification, recognition, and tracking, also serve as a viable sensor type for AD. Yet, none of the primary studies in our review employed cameras for this purpose. 
Nevertheless, the integration of general-purpose or specialized cameras in AD systems could provide valuable insights and broaden the scope of detectable anomalies. For instance, the use of event-based cameras for vibration analysis is a promising application, as they provide superior spatio-temporal resolution in comparison to traditional sensors~\cite{lai2020full}. 

\subsection{What are the types of Machine Learning techniques used for anomaly detection in industrial machinery?}
\label{subsec_RQ3}

There are two critical stages in the development of an effective AD technique for industrial machinery: data preprocessing and the application of AI algorithms. 
Both stages are equally important and play fundamental roles in the ability to accurately and timely detect anomalies. In the following subsections,  we provide a detailed description of the preprocessing techniques and algorithms used in the selected primary studies for AD in industrial machinery.

\subsubsection{Preprocessing techniques}

Preprocessing techniques play a crucial role in the field of ML, helping to clean and prepare data before it is fed into algorithms for analysis. The process of preprocessing involves a range of techniques such as data cleaning, normalization, feature selection, and dimensionality reduction, among others~\cite{geron2022hands}. These techniques are critical for improving the quality of data, reducing computational complexity, and enhancing the performance of ML models.
According to the results obtained in the analysis of the selected primary studies, it was found that approximately 67\% of the studies presented some preprocessing technique applied to the data used in the research. The techniques involved in preprocessing could range from feature selection to dimensionality reduction, and data transformations aimed at enhancing the quality of the input data. This reinforces the critical role of preprocessing in AD for industrial machinery, as the use of appropriate techniques can significantly improve the efficiency and accuracy of the detection process.

\begin{figure}[!htbp]
    \centering
    \includegraphics[width=\columnwidth]{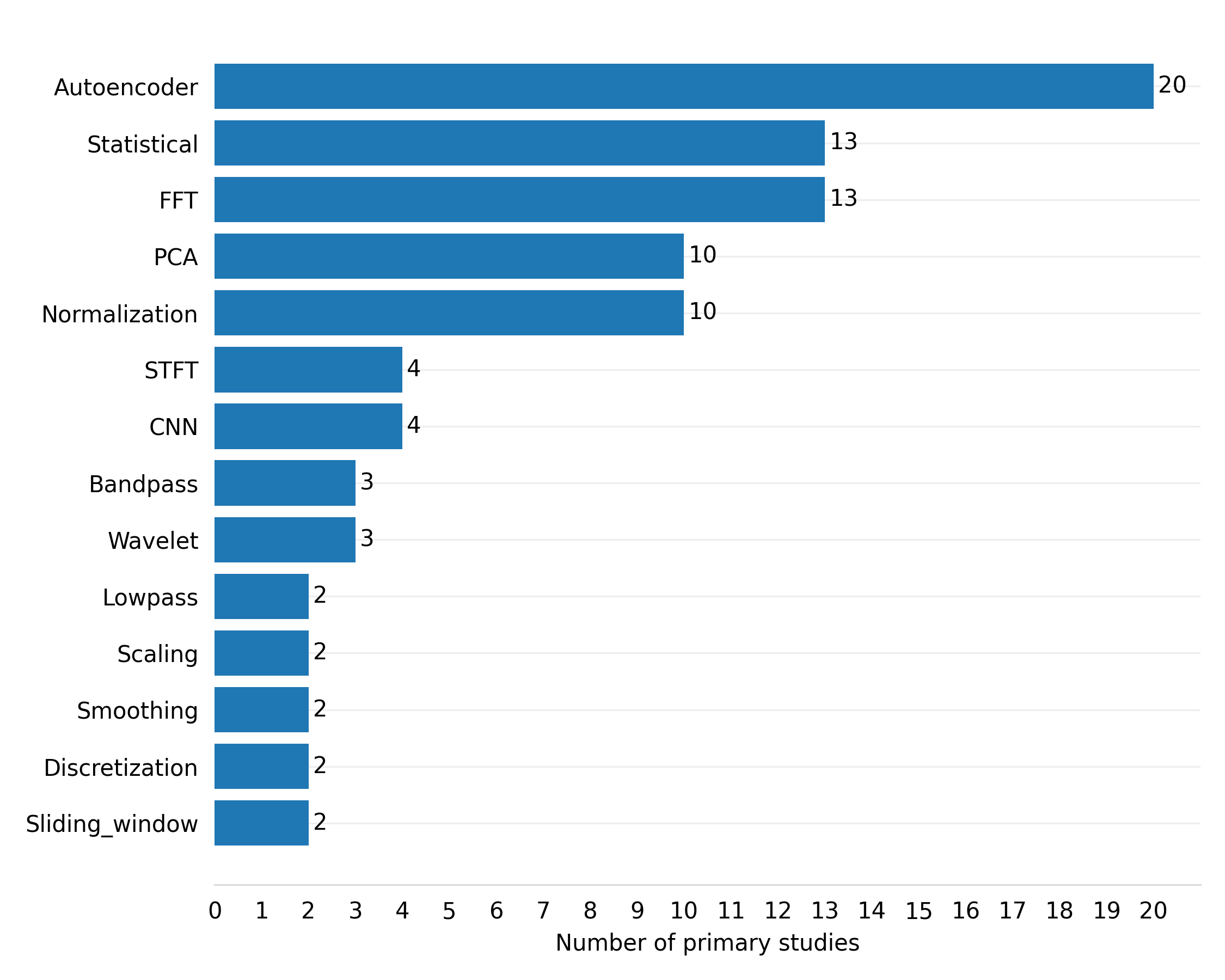}
    \caption{Number of primary studies by preprocessing techniques.}
    \label{fig:pre-processamento}
\end{figure}

Figure~\ref{fig:pre-processamento} illustrates the most frequently employed preprocessing techniques in the selected primary studies. 
The most commonly adopted data preprocessing technique was the Autoencoder, which is a powerful technique for dimensionality reduction \cite{baldi2012autoencoders}. Autoencoders use neural networks to learn a compressed representation of the input data. This technique can be especially useful for nonlinear and high-dimensional data,  where traditional methods such as Principal Component Analysis (PCA) may not be as effective~\cite{wickramasinghe2021resnet}.  
Other techniques were categorized together because they served a similar purpose. For instance, statistical methods such as mean, median, standard deviation, and variance were used to measure the central tendency and variability of the data and were grouped together. This group is referred as ``Statistical'' and is ranked as the second most commonly utilized technique for preprocessing. These statistical methods enable researchers to gain insights into the distribution of data, identify outliers, and detect trends or patterns~\cite{ali2016basic}.

Fast Fourier Transform (FFT), another frequently used technique, and Short-Time Fourier Transform (STFT) both operate in the frequency domain, enabling the time-based analysis of signals. However, FFT is better suited for stationary signals, while STFT is more appropriate for non-stationary signals~\cite{aggarwal2011noise}. Among the identified techniques, FFT was the most widely used, appearing in 13 studies, while STFT was used in 4 studies.
Another highly adopted technique for data preprocessing was PCA (Principal Component Analysis). It identifies the most important variables in a dataset and reduces the number of dimensions while preserving the most significant information. By transforming the data into a new coordinate system that maximizes the variance of the data, PCA helps to eliminate redundant or noisy features that may hinder the analysis~\cite{groth2013principal}. Normalization is another widely adopted preprocessing technique that is commonly found in primary studies.  Its importance lies in transforming data into a standardized scale, allowing for the easy comparison and analysis of data from diverse sources. Additionally, it helps mitigate any inherent bias in the data that may arise due to differences in units of measurement or data collection procedures~\cite{smolinska2014current}. From the selected primary studies, it was found that PCA was utilized in 10 studies, and normalization was employed in an equal number of studies.
 
Another preprocessing technique that was moderately used in the primary studies was Convolutional Neural Network (CNN). It is a deep learning (DL) architecture that has achieved remarkable success in image and signal processing tasks, thanks to its ability to learn hierarchical representations of the input data. Besides its primary application in image and signal processing, CNN can also be used for feature extraction in various applications~\cite{jogin2018feature}.

The popularity of these techniques highlights their importance in data preprocessing and analysis, as they enable researchers to reduce the dimensionality of complex data and extract relevant features that can improve the accuracy and efficiency of subsequent analysis.

While less commonly used, several other preprocessing techniques were also found in the selected studies. These include bandpass and low-pass filters~\cite{conradi2021anomaly, jian2021initial, li2020lifelong, watanabe2020anomaly, gruner2020evaluation}, wavelet transformations~\cite{assafo2021topsis, ying2021hybrid, li2020lifelong}, sliding window techniques~\cite{wu2019lstm, zhou2022contrastive}, and others. However, it is worth mentioning that the selection of the most appropriate preprocessing techniques depends on the nature and specificity of the data being analyzed, and researchers must consider the advantages and limitations of each technique before making a decision.

\begin{figure}[!htbp]
    \centering
    \includegraphics[width=\columnwidth]{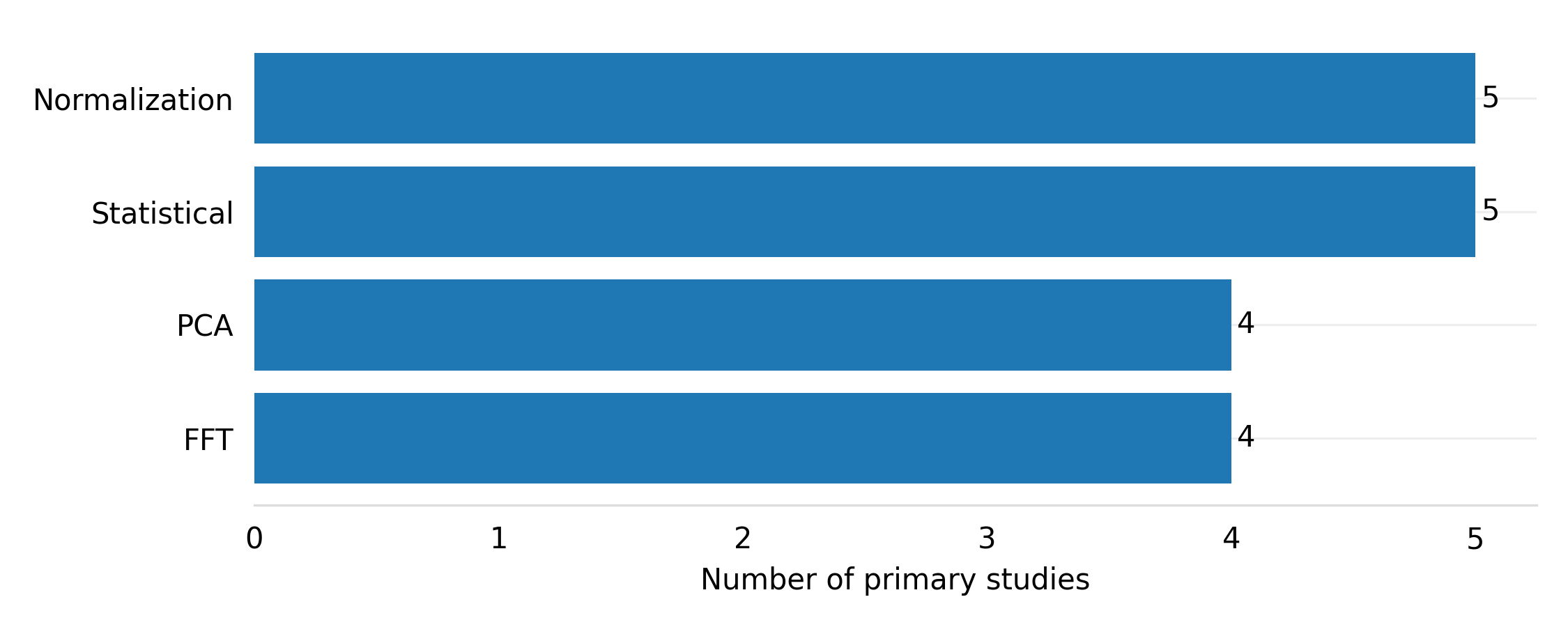}
    \caption{Number of primary studies by preprocessing techniques filtered by autoencoder.}
    \label{fig:Number_of_primary_studies_by_preprocessing_filter_autoencoder}
\end{figure}

Additionally, some studies combined preprocessing techniques to try to improve AD. For instance, a notable example is the utilization of autoencoders in conjunction with other preprocessing methods. Among the studies that employed autoencoders, five studies used normalization or statistics in conjunction with autoencoder, while four studies combined FFT or PCA with autoencoder, as shown in Figure ~\ref{fig:Number_of_primary_studies_by_preprocessing_filter_autoencoder}. These results suggest that using multiple preprocessing techniques together can be an effective way to enhance AD accuracy.

\subsubsection{Algorithms}

\begin{figure}[!htbp]
    \centering
    \includegraphics[width=\columnwidth]{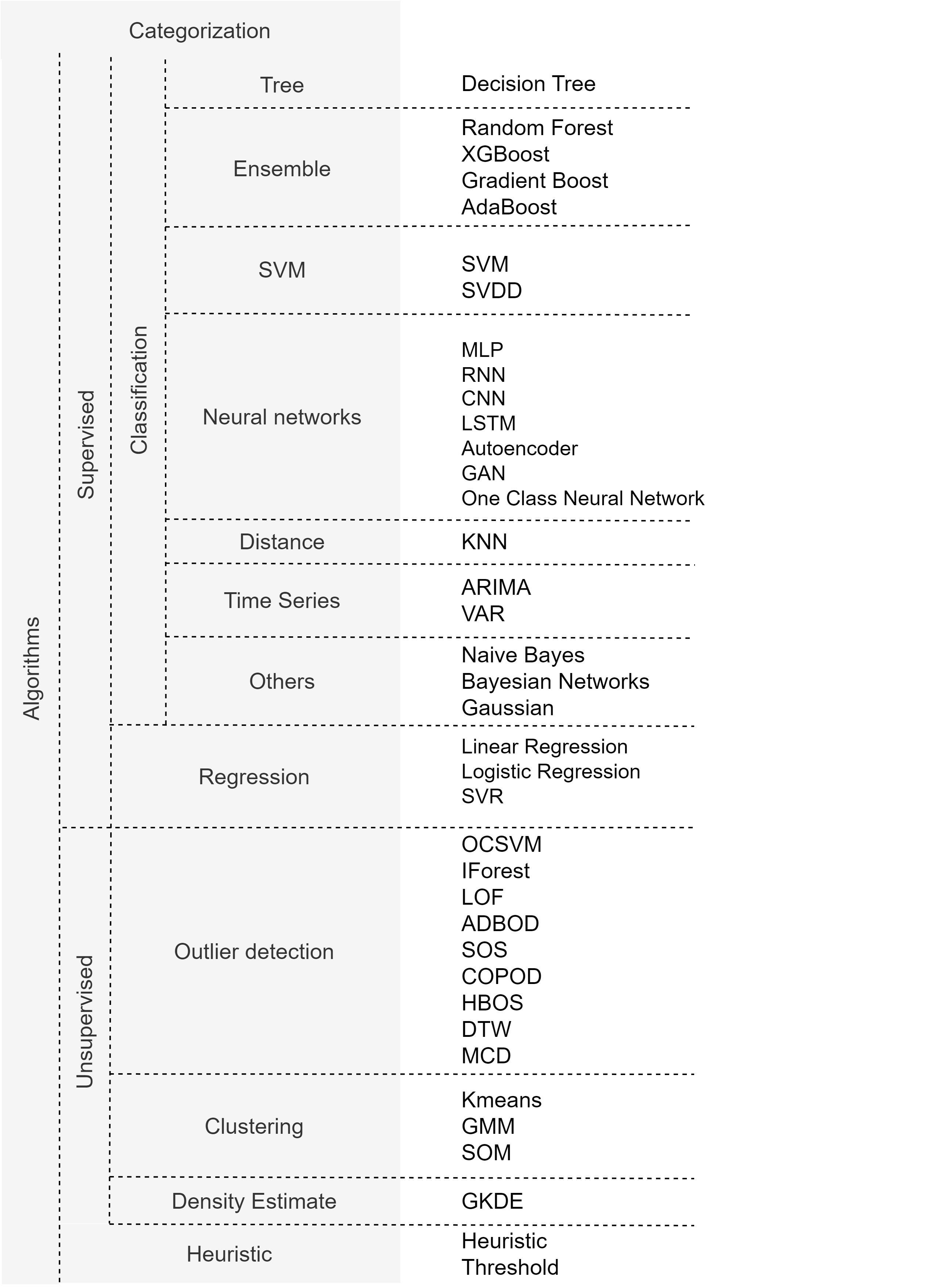}
    \caption{Algorithms categorization.}
    \label{fig:algorithms-categorization}
\end{figure}

Various ML-based algorithms have been employed to detect anomalies in industrial machinery. Figure~\ref{fig:algorithms-categorization} presents these algorithms grouped into different levels of categorization. Vertical categories indicate further subdivisions, while horizontal categories represent the final categorization. The initial grouping divides the algorithms into three categories: Supervised, Unsupervised, and Heuristics.
Supervised models were divided into two categories: Classification and Regression. Classification models were subdivided into several types: tree-based models, ensemble models, SVM, neural networks, distance-based models, time-series models, and others. Regression models, in turn, were the final categorization. The unsupervised models, on the other hand, were divided into three final categories: outlier detection, clustering, and density estimation. Another separate final grouping was the Heuristic.

\begin{figure}[!htbp]
    \centering
    \includegraphics[width=\columnwidth]{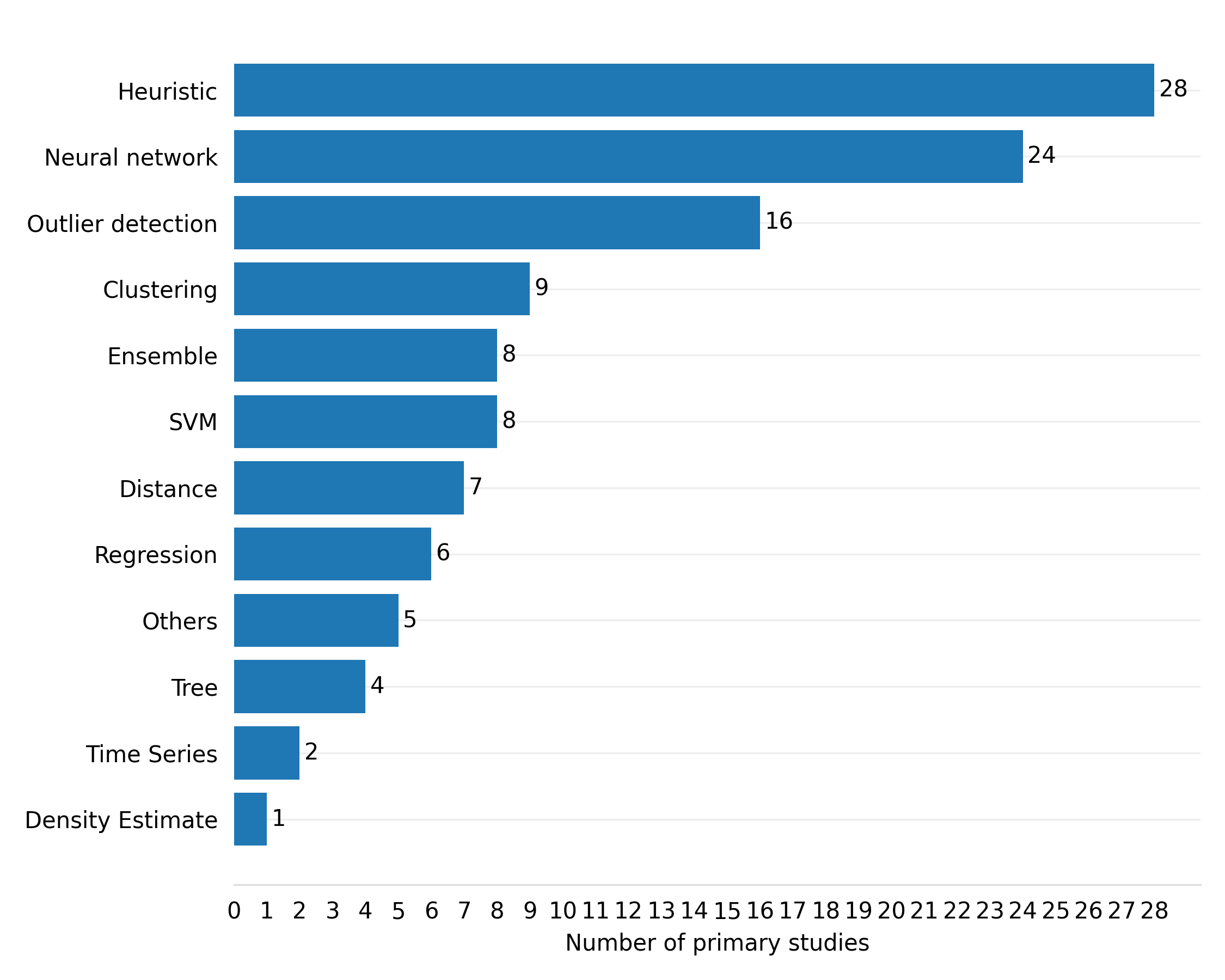}
    \caption{Number of primary studies based on algorithms categorization.}
    \label{fig:Number_of_primary_studies_by_Algorithms_Categorization_level_3}
\end{figure}

Figure \ref{fig:Number_of_primary_studies_by_Algorithms_Categorization_level_3} shows the number of primary studies that used each final algorithm categorization. There are 12 final categorizations in total. This includes seven groups for classification, one for regression, three for unsupervised learning, and one for heuristic. It is important to note that a single study might use more than one technique.
The results of the studies demonstrate the effectiveness of various algorithms and approaches for identifying anomalies in industrial machinery. Heuristic-based approaches were the most common, used in 28 studies. Neural network methods were also widely used, found in 24 primary studies. Outlier detection was another common method for finding anomalies in sensor data, used in 16 primary studies. On the other hand, time-series-based and density estimate approaches were less frequently used, with only 2~\cite{abbracciavento2021anomaly, liu2021online} and 1~\cite{yanabe2020anomaly} primary studies, respectively.

Based on the categorizations that were most prominent in Figure \ref{fig:Number_of_primary_studies_by_Algorithms_Categorization_level_3}, we are going to analyze which techniques are frequently used in one of these categories. We start with the methods related to neural networks. After that, we address the techniques used for outlier detection. Finally, we analyze the techniques used in the most common categorization, which is Heuristic.

\begin{figure}[!htbp]
    \centering
    \includegraphics[width=0.9\columnwidth]{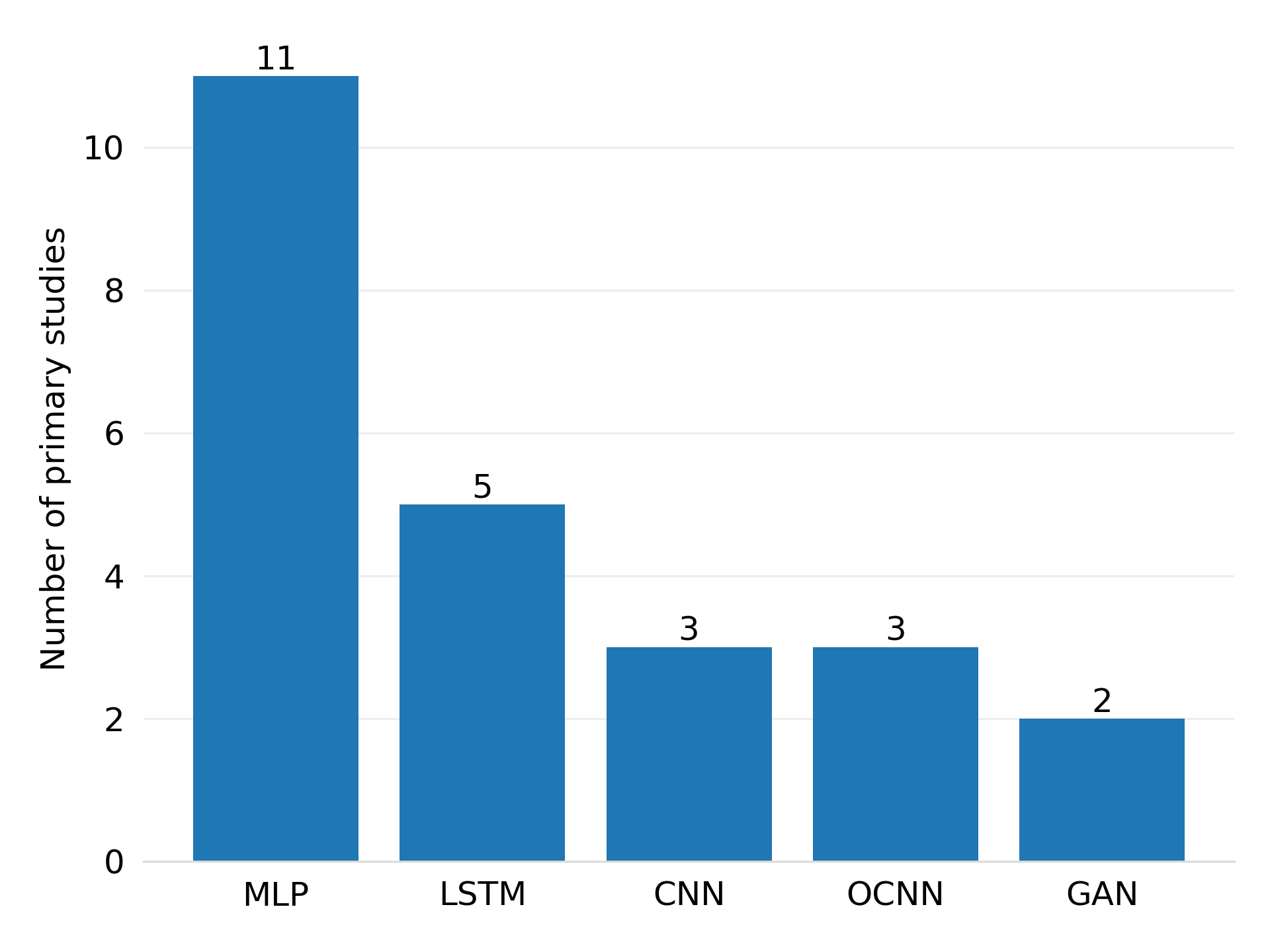}
    \caption{Number of primary studies by neural network category.}
    \label{fig:Number_of_primary_studies_by_algorithms_filter_Neural_network}
\end{figure}

Figure \ref{fig:Number_of_primary_studies_by_algorithms_filter_Neural_network} illustrates the most frequently employed neural network techniques in the primary studies related to ML-based AD. 
Multilayer perceptron (MLP) was the most widely used technique, with 11 primary studies, which involves multiple layers of artificial neurons used to recognize patterns in data. 
Long short-term memory (LSTM) was the second most popular technique with five primary studies \cite{wu2019lstm, wang2021early, canizo2019multi, wielgosz2019mapping, de2022long}, which is a type of recurrent neural network (RNN) architecture designed to remember past data over time. 
CNN was used in three primary studies \cite{ying2021hybrid, yun2020development, meire2019comparison}, which is a type of feed-forward neural network that was also employed in the preprocessing step, as mentioned previously.
One-class neural network (OCNN) was employed in three primary studies \cite{jiang2022multiscale, bose2019adepos, boons2021edge}, which is a neural network architecture specifically designed for AD problems. Finally, Generative Adversarial Networks (GANs) were employed in two primary studies \cite{tagawa2021acoustic, liang2021robust}, which is a type of neural network that involves two models trained simultaneously, where one model generates samples, and the other model evaluates their authenticity ~\cite{creswell2018generative}. The less frequent implementation of GANs, relative to other DL architectures, may not be a reflection of its applicability for anomaly detection, but rather of its relative novelty in the field, which suggest that this approach could be explored further. 

\begin{figure}[!htbp]
    \centering
    \includegraphics[width=\columnwidth]{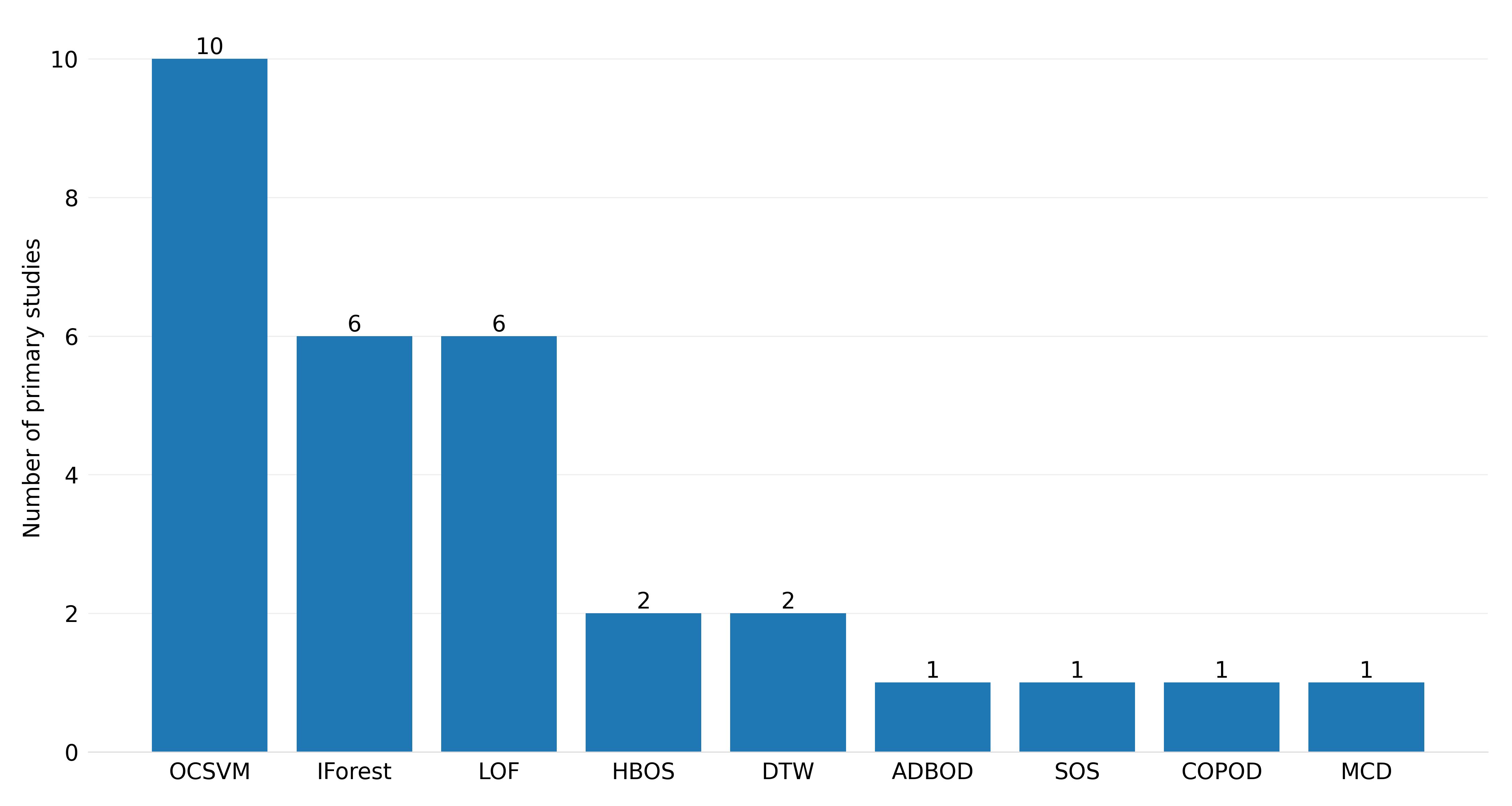}
    \caption{Number of primary studies by outlier detection category.}
    \label{fig:Number_of_primary_studies_by_algorithms_filter_anomaly_detection.}
\end{figure}

The outlier detection category has also received significant attention in the literature. Figure \ref{fig:Number_of_primary_studies_by_algorithms_filter_anomaly_detection.} shows the specific techniques employed in this category. The most commonly used techniques among the primary studies were One-class support vector machine (OCSVM) with 10 studies, followed by Isolation Forest (Iforest) and Local Outlier Factor (LOF), with 6 studies each. 
OCSVM is a well-known method for outlier detection that separates inliers from outliers by identifying a boundary around the inliers~\cite{erfani2016high}. Iforest is a tree-based algorithm that isolates outliers by constructing separation trees ~\cite{liu2008isolation, cheng2019outlier}. LOF is a density-based method that measures the local deviation of a data point with respect to its neighbors~\cite{cheng2019outlier}. Although other techniques such as Histogram-based Outlier Score (HBOS), Dynamic Time Warping (DTW), and Angle-Based Outlier Detection (ADBOD) were also adopted, their usage was relatively limited in comparison to the previously mentioned approaches.

In the primary studies that utilized heuristics to detect anomalies, the most prevalent approach involved setting a threshold to identify values that significantly deviated from normal behavior. 
This approach, known as rule-based AD, is widely used in industrial applications. However, using a fixed value or percentage to set the threshold can lead to false positives or false negatives. To enhance detection accuracy, this technique is typically combined with preprocessing steps.

\begin{figure}[!htbp]
    \centering
    \includegraphics[width=\columnwidth]{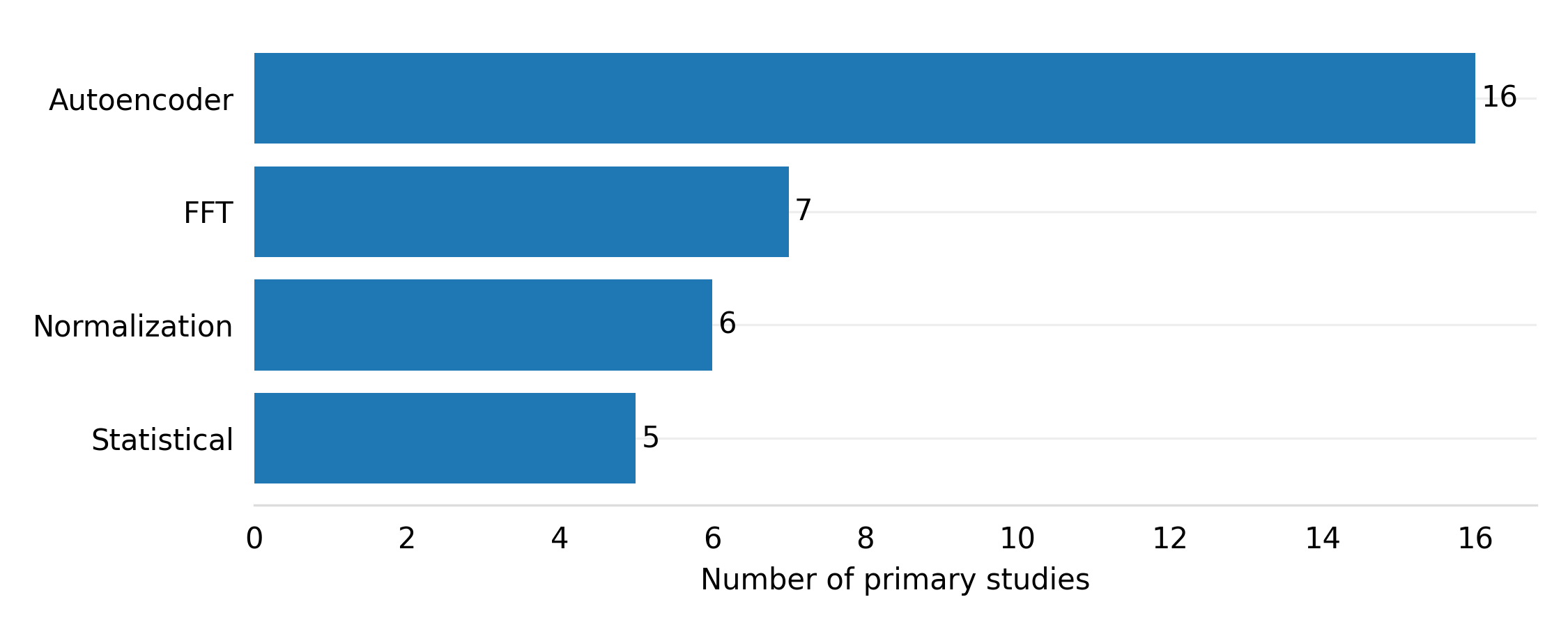}
    \caption{Number of primary studies categorized by preprocessing techniques and filtered by the heuristic approach.}
    \label{fig:Number_of_primary_studies_by_preprocessing_filter_heuristic}
\end{figure}

Figure \ref{fig:Number_of_primary_studies_by_preprocessing_filter_heuristic} illustrates the most commonly used preprocessing techniques in conjunction with this approach.
Among the various preprocessing methods, the autoencoder technique was the most commonly utilized, with 16 instances.  Several studies have demonstrated the integration of this technique with threshold-based rule-based AD, where the threshold is established based on the autoencoder reconstruction error. 
Other techniques such as FFT, normalization, and statistics were also widely used to assist in diagnosis using thresholds. These findings demonstrate that the heuristics-based approach for AD is promising, but requires careful consideration  in selecting an appropriate preprocessing technique and defining the threshold for each specific application.

\subsection{How is the anomaly detection method  computed and evaluated?}
\label{subsec_RQ4}

\begin{figure}[!htbp]
    \centering
    \includegraphics[width=\columnwidth]{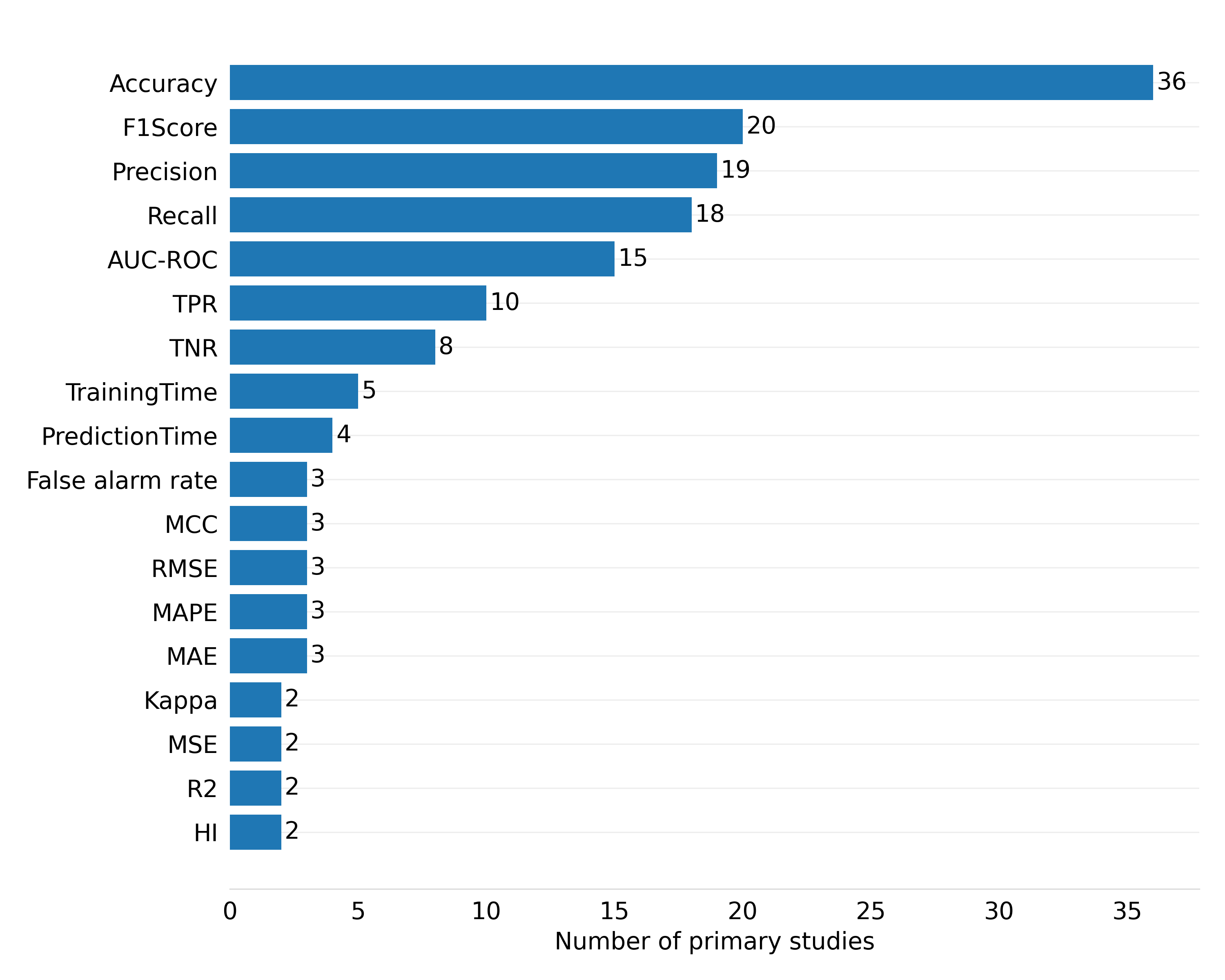}
    \caption{Number of primary studies by metrics.}
    \label{fig:metrics}
\end{figure}

The choice of an appropriate evaluation method is crucial when aiming to find the optimal classifier for a given problem, whether it involves classification, regression, or clustering ~\cite{hossin2015review}. 
Evaluating AD models can be done using various metrics that align with specific tasks. In our primary studies, we identified several metrics, as shown in Figure~\ref{fig:metrics}.
The metric \textit{Accuracy} was the most commonly used in the analyzed studies, appearing in 36 instances. It was closely followed by \textit{F1-score}, \textit{Precision} \textit{Recall}, which appeared in 20, 19, and 18 studies, respectively. These metrics are derived from the confusion matrix, a tabular representation that summarizes the model's performance by indicating correct and incorrect classifications in terms of true positives, true negatives, false positives, and false negatives.

The Area Under the Receiver Operating Characteristic Curve \textit{(AUC-ROC)} is a widely utilized metric in binary classification models. It represents the relationship between the true positive rate (TPR) and the false positive rate (FPR) at different threshold levels. Our analysis identified this specific metric in 15 primary studies.
Additionally, the \textit{TPR} and \textit{FPR} metrics, which are foundational for calculating the ROC curve, were found in 10 and 8 studies, respectively. Consequently, it can be deduced that binary classification is the most frequently performed and evaluated task within the context of AD. 

Other metrics employed in binary classification include \textit{Matthews Correlation Coefficient (MCC)}, frequently used metric for unbalanced classes, which combines true positive, true negative, false positive, and false negative rates to evaluate overall model quality, used in 3 studies \cite{ds2022comparative, zufle2022workflow, liang2021robust}; \textit{False alarm rate}, also known as FPR, which quantifies the proportion of negatives mistakenly classified as positives, also used in 3 studies \cite{lughofer2020line, assafo2021topsis, zabinski2021fpga}; and \textit{Kappa}, which measures the level of agreement between observers or classifiers in a classification problem, present in 2 studies ~\cite{ds2022comparative, kilian2022vibration}.

In addition to classification-focused metrics, there are also metrics that specifically target regression models. \textit{Mean Absolute Error (MAE)}, \textit{Mean Absolute Percentage Error (MAPE)}, \textit{Root Mean Square Error (RMSE)}, \textit{Mean Squared Error (MSE)}, and \textit{Coefficient of Determination (R2)} are a few of these. By quantifying the discrepancy between predicted values and actual values, these metrics seek to assess the precision of numerical predictions and offer information about how well the model fits the regression data. MAE, MAPE, and RMSE were the most commonly employed regression metrics, appearing in 3 studies each~\cite{ayvaz2021predictive, kim2019online, yanabe2020anomaly, song2019anomaly, apostol2021change}. R2 and MSE were present in 2 studies~\cite{ayvaz2021predictive, ying2021hybrid, mostafavi2021novel}.

Most regression-based models are commonly applied in solving Prognostics and Health Management (PHM) problems, as demonstrated in the works by~\cite{zhao2022lithium, berghout2022systematic, siahpour2022novel}. These studies primarily focus on predicting the remaining lifetime of systems using monitoring data as input. An alternative approach to tackle this scenario is through the application of a time series approach, as observed in the works by~\cite{song2022novel, lin2020battery}. However, this approach often emphasizes predicting the future state of systems rather than real-time state identification. Within this context, the \textit{Health Indicator (HI)} metric, as employed in Zhai \textit{et al.} (2021)~\cite{zhai2021enabling} and Guo \textit{et al.} (2022)~\cite{guo2022unsupervised}, aims to evaluate the health status of monitored equipment.

Metrics concerning time performance, such as training time and prediction time, were also identified in 5 \cite{di2022anomalous, brito2022explainable, meire2019comparison, yan2023hybrid, mostafavi2021novel} and 4 \cite{di2022anomalous, park2021design, meire2019comparison, yan2023hybrid} studies, respectively. These metrics are not unique to any specific model type but serve to assess the efficiency and duration of model training and prediction generation. They offer valuable insights into the scalability and practical feasibility of implementing the model, considering computational resources and processing time. These metrics are crucial for optimizing resource allocation, determining model suitability for real-time applications, and evaluating overall system performance in time-sensitive scenarios.

Finally, it is clear that there  is a lack of research in the literature on the assessment of inference time, i.e. the time the proposed approach takes to process the data stream and classify it as normal or anomalous. Importantly, only 4 of the 84 works in this mapping report the inference time of the proposed model. This emphasizes the necessity for additional study to address this issue and develop a thorough grasp of the computational needs and applicability of AD methods in real-time. By examining prediction time alongside other performance metrics, we can improve decision-making processes and identify models that strike a balance between accuracy and efficiency in time-critical domains.

\section{Open Challenges}
\label{sec_OpenChallenges}

While AD techniques for industry 4.0 are gaining attention in recent years, several open challenges have been identified by this SM. This section highlights these challenges and provides insights into potential areas for future research.

\subsection{Anomaly detection on the edge} The rise of edge computing offers promising potential for real-time AD, particularly for IIoT sensors. However, edge devices are constrained by limited resources, such as processing power and memory, making complex anomaly detection tasks challenging to execute on these devices. To address this challenge, there is a need to develop suitably optimized AD algorithms that can navigate the trade-off between precision and computational resources. Multiobjective optimization algorithms can be used to finetune solutions under simultaneous conflicting objectives, such as power consumption, accuracy and inference speed \cite{emmerich2018tutorial}. Additionally, spiking neural networks and neuromorphic computing are emerging technologies that can be used for real-time anomaly detection with significantly lower computational resources than traditional approaches~\cite{bauer2019real}.

\subsection{Usage of low-cost and off-the-shelf sensors} 
The adoption of IIoT devices with low-cost sensors could facilitate a wider implementation of predictive maintenance in the industry. However, a prominent challenge lies in the quality and reliability of the data collected from these sensors. Low-cost sensors may exhibit higher noise levels, lower accuracy, and reduced stability over time compared to their more expensive counterparts. These factors can significantly impact the performance of AD algorithms, resulting in a higher rate of false positives. Consequently, AD algorithms would require additional adjustments, such as implementing an automatic update routine to compensate for the lower reliability and potential sensory drift over time.

\subsection{Identification of the anomaly source} 
While AD techniques can indicate the occurrence of an anomaly based on normal operation data alone, fault diagnosis remains a challenging next step. Fault diagnosis is crucial for understanding the root cause of the anomaly and taking appropriate actions.
With the emergence of  XAI, there is an increasing demand for models that not only detect anomalies but also provide interpretable information about their likely causes. This explanation of the potential causes of the anomaly is even more valuable in environments where downtime is critical, as it enables operators to make faster decisions.

\subsection{Lack of fault data} 
In industrial settings, it is often the case that there is an abundance of non-anomalous data, representing healthy machinery, while the amount of anomalous data is relatively small or even non-existent. This represents a significant challenge when it comes to using supervised learning approaches, as these methods heavily rely on labeled anomalous data for training and generalization of the models. Therefore, addressing this challenge becomes crucial.
One possible solution to tackle the lack of anomalous data is to leverage a combination of simulation and experimental data. By utilizing simulated scenarios and conducting controlled experiments, it becomes possible to generate synthetic anomalous data. This synthetic data can then be used for the initial validation and fine-tuning of the AD algorithm in a more controlled and manageable environment.

\subsection{Integration with existing infrastructure}
One of the key challenges in AD is the integration of AD systems with the pre-existing infrastructure in industrial environments. Industrial systems often have well-established data collection mechanisms and processes in place. Therefore, it is essential for AD solutions to be compatible with these existing systems and seamlessly integrate into the infrastructure without causing disruptions or requiring significant changes. The integration aspect should be a primary consideration in research studies, and future works should emphasize how their approaches can effectively adapt to the already consolidated industrial environments, ensuring smooth incorporation without major interruptions or modifications.

\subsection{Retraining of ML models}
The retraining of ML models is a crucial aspect that requires attention and careful discussion regarding its adaptability to the evolution of industries, changes in the environment and new data collected.
Ensuring the continued effectiveness of ML models in detecting anomalies depends on their ability to adapt to changes that occur over time in industries. As industries undergo operational changes, these models need to be capable of adjusting and learning from the new environment. Furthermore, the inclusion of newly collected data and exposure to new failures can enable the models to identify anomalies with greater precision.

%\subsection{Practical applications}
%\textcolor{black}{Notably, most of the mapped studies rely on existing datasets for benchmarking. Thus, very few studied provide a tangible application of IoT devices and anomaly detection in an industrial setting. As an exception, Ayvaz and Alpay~\cite{ayvaz2021predictive} propose a predictive maintenance system for production lines. By using real-world manufacturing IoT data, the system can provide indicators of imminent failures and mitigate potential production stops, with models of Random Forest and XGBoost demonstrating better performance in comparative evaluations. Also, Kim and Heo~\cite{kim2022anomaly} use data from IoT sensors and apply various ML algorithms for anomaly detection for a hydraulic system. By extracting a total of 2335 features, the study is able to identify stable and unstable conditions in hydraulic system components.
%}

\section{Limitations of the Mapping Study}
\label{sec_limitations}

Risks and restrictions apply to systematic mappings like this one~\cite{petersen2008systematic}. In this section, we address the most frequent restrictions encountered during the review process and provide our solutions to overcome them.

\emph{Formulation of research questions:} The development of research questions is essential for directing the review process. However, if the questions are not carefully designed or if they lack specificity, it is possible to mistakenly exclude pertinent studies or ignore crucial components. By carefully crafting and fine-tuning the research questions through consultations with authors and outside experts, we were able to lessen this constraint.

\emph{Searching:} Despite using a thorough search method, it is probable that some pertinent studies were missed. This can be as a result of restrictions on the databases that were chosen, limitations on language, or the exclusion of specific publishing categories. We maintained the same terms in our search strings while modifying them for each digital database to lessen this restriction.

\emph{Misclassification or errors in data extraction:} These terms allude to the potential for various reviewers to interpret the data from research in different ways. Although we used our judgment to categorize the studies, there is still a chance that we did so incorrectly. Multiple author-researchers were involved in the classification process to help alleviate this possible problem, and any disagreements were settled by consensus discussions.

We aim to provide a comprehensive and unbiased evaluation of the literature within the scope of this review, acknowledging these limitations and employing appropriate methodologies.

\section{Conclusions and next steps}
\label{sec_final}

This SM focused on the topic of AD in industrial machinery using IoT devices and ML algorithms. The study aimed to fill the gap in existing research by providing a comprehensive review of current methodologies, synthesis of evidence, areas of application, research questions and future challenges in this domain.
Through the analysis of 84 studies dating from 2016 to 2023, several important findings emerged. First, it was observed that the types of machinery most commonly monitored include milling and cutting tools, hydraulic systems, and bearings. The analysis suggests that these types of machinery are often subjected to conditions that can lead to anomalies, such as wear and tear in milling and cutting tools, fluctuations in temperature and pressure in hydraulic systems, and heat-related failures in bearings. 

Second, a wide range of sensors were employed to detect anomalies. The primary works often did not detail specific characteristics of the sensors, but we were able to highlight the most used sensors, such as vibration and temperature. A combination of these sensors is observed as a common strategy to efficiently detect and monitor anomalies in machines.
Third, several ML techniques have been used for AD, including supervised, unsupervised and heuristic methods. In addition to highlighting preprocessing techniques, we were able to highlight the most used and also those that normally worked together. Combination of these techniques, such as FFT and autoencoders, are found to be effective to enhance AD accuracy. 

The review also identified several open challenges in the field of AD. \textcolor{black}{These challenges include detecting anomalies at the edge using off-the-shelf and low-cost sensors, enabling faster and cost-effective upgrades to existing infrastructure.  
Additionally, beyond detection, identifying the source of anomalies can offer valuable insights for predictive maintenance.  Furthermore, addressing the scarcity of fault data and retraining ML models to adapt to evolving industrial environments are ongoing challenges in this field.} In summary, this SM clarified the state of the art of AD in industrial machinery using IoT devices and ML. It identified key research areas, highlighted current methodologies, and outlined open challenges. By addressing these challenges, we aim to contribute to the development of more accurate, efficient, and interpretable AD systems, thereby enhancing industrial productivity, safety, and efficiency in the era of smart manufacturing

For future work, our intention is to develop an anomaly detection solution based on low-cost and off-the-shelf components for rotating machinery. Additionally, we aim to focus on developing anomaly detection on embedded devices, with a priority on low power consumption and local processing of most of the data. As our research progresses, we plan to make newly created anomaly datasets publicly accessible, contributing to the broader research community's resources. \textcolor{black}{Finally, future literature reviews could provide an in depth analysis of existing industrial dataset for AD and predictive maintenance.}

\section*{AUTHOR CONTRIBUTIONS}

Sérgio F. Chevtchenko: Writing - original draft, conceptualization, methodology, investigation, data curation, and formal analysis. 
Elisson da Silva Rocha: Methodology, Writing - original draft, data curation, and formal analysis. 
Monalisa Cristina Moura dos Santos: Writing - original draft and review, conceptualization, methodology, and data curation.
Diego Moura Vieira: Writing - original draft and review, conceptualization, methodology, and data curation.
Ricardo Lins Mota: Writing - original draft and review, conceptualization, methodology, and data curation.
Ermeson Carneiro de Andrade: Methodology, Writing - review and editing, Supervision.
Danilo Ricardo Barbosa de Araújo: Methodology, Writing - review and editing, Supervision. 

\bibliographystyle{unsrt}
\bibliography{PANM_paper}

\begin{thebibliography}{100}

\bibitem{lee2014service}
Jay Lee, Hung-An Kao, and Shanhu Yang.
\newblock Service innovation and smart analytics for industry 4.0 and big data environment.
\newblock {\em Procedia cirp}, 16:3--8, 2014.

\bibitem{kaupp2021context}
Lukas Kaupp, Heiko Webert, Kawa Nazemi, Bernhard Humm, and Stephan Simons.
\newblock Context: An industry 4.0 dataset of contextual faults in a smart factory.
\newblock {\em Procedia Computer Science}, 180:492--501, 2021.

\bibitem{chandola2009anomaly}
Varun Chandola, Arindam Banerjee, and Vipin Kumar.
\newblock Anomaly detection: A survey.
\newblock {\em ACM computing surveys (CSUR)}, 41(3):1--58, 2009.

\bibitem{chatterjee2022iot}
Ayan Chatterjee and Bestoun~S Ahmed.
\newblock Iot anomaly detection methods and applications: A survey.
\newblock {\em Internet of Things}, 19:100568, 2022.

\bibitem{huch2018machine}
Fabian Huch, Mojdeh Golagha, Ana Petrovska, and Alexander Krauss.
\newblock Machine learning-based run-time anomaly detection in software systems: An industrial evaluation.
\newblock In {\em 2018 IEEE Workshop on Machine Learning Techniques for Software Quality Evaluation (MaLTeSQuE)}, pages 13--18. IEEE, 2018.

\bibitem{ahmed2016survey}
Mohiuddin Ahmed, Abdun~Naser Mahmood, and Jiankun Hu.
\newblock A survey of network anomaly detection techniques.
\newblock {\em Journal of Network and Computer Applications}, 60:19--31, 2016.

\bibitem{eltanbouly2020machine}
Sohaila Eltanbouly, May Bashendy, Noora AlNaimi, Zina Chkirbene, and Aiman Erbad.
\newblock Machine learning techniques for network anomaly detection: A survey.
\newblock In {\em 2020 IEEE International Conference on Informatics, IoT, and Enabling Technologies (ICIoT)}, pages 156--162. IEEE, 2020.

\bibitem{wang2021machine}
Song Wang, Juan~Fernando Balarezo, Sithamparanathan Kandeepan, Akram Al-Hourani, Karina~Gomez Chavez, and Benjamin Rubinstein.
\newblock Machine learning in network anomaly detection: A survey.
\newblock {\em IEEE Access}, 9:152379--152396, 2021.

\bibitem{dalzochio2020machine}
Jovani Dalzochio, Rafael Kunst, Edison Pignaton, Alecio Binotto, Srijnan Sanyal, Jose Favilla, and Jorge Barbosa.
\newblock Machine learning and reasoning for predictive maintenance in industry 4.0: Current status and challenges.
\newblock {\em Computers in Industry}, 123:103298, 2020.

\bibitem{zonta2020predictive}
Tiago Zonta, Cristiano~Andr{\'e} Da~Costa, Rodrigo da~Rosa~Righi, Miromar~Jose de~Lima, Eduardo~Silveira da~Trindade, and Guann~Pyng Li.
\newblock Predictive maintenance in the industry 4.0: A systematic literature review.
\newblock {\em Computers \& Industrial Engineering}, 150:106889, 2020.

\bibitem{napoleao2017practical}
Bianca Napole{\~a}o, Katia~Romero Felizardo, {\'E}rica~Ferreira de~Souza, and Nandamudi~L Vijaykumar.
\newblock Practical similarities and differences between systematic literature reviews and systematic mappings: a tertiary study.
\newblock In {\em SEKE}, volume 2017, pages 85--90, 2017.

\bibitem{bandyopadhyay2011internet}
Debasis Bandyopadhyay and Jaydip Sen.
\newblock Internet of things: Applications and challenges in technology and standardization.
\newblock {\em Wireless personal communications}, 58:49--69, 2011.

\bibitem{lu2017industry}
Yang Lu.
\newblock Industry 4.0: A survey on technologies, applications and open research issues.
\newblock {\em Journal of industrial information integration}, 6:1--10, 2017.

\bibitem{cunha2021upgrading}
Jo{\~a}o Cunha, Nelson Batista, Carlos Cardeira, and Rui Melicio.
\newblock Upgrading a legacy manufacturing cell to iot.
\newblock {\em Journal of Sensor and Actuator Networks}, 10(4):65, 2021.

\bibitem{ayvaz2021predictive}
Serkan Ayvaz and Koray Alpay.
\newblock Predictive maintenance system for production lines in manufacturing: A machine learning approach using iot data in real-time.
\newblock {\em Expert Systems with Applications}, 173:114598, 2021.

\bibitem{sisinni2018industrial}
Emiliano Sisinni, Abusayeed Saifullah, Song Han, Ulf Jennehag, and Mikael Gidlund.
\newblock Industrial internet of things: Challenges, opportunities, and directions.
\newblock {\em IEEE transactions on industrial informatics}, 14(11):4724--4734, 2018.

\bibitem{angelopoulos2019tackling}
Angelos Angelopoulos, Emmanouel~T Michailidis, Nikolaos Nomikos, Panagiotis Trakadas, Antonis Hatziefremidis, Stamatis Voliotis, and Theodore Zahariadis.
\newblock Tackling faults in the industry 4.0 era—a survey of machine-learning solutions and key aspects.
\newblock {\em Sensors}, 20(1):109, 2019.

\bibitem{wuest2016machine}
Thorsten Wuest, Daniel Weimer, Christopher Irgens, and Klaus-Dieter Thoben.
\newblock Machine learning in manufacturing: advantages, challenges, and applications.
\newblock {\em Production \& Manufacturing Research}, 4(1):23--45, 2016.

\bibitem{kumar2022intrusion}
Ravinder Kumar, Amita Malik, and Virender Ranga.
\newblock An intellectual intrusion detection system using hybrid hunger games search and remora optimization algorithm for iot wireless networks.
\newblock {\em Knowledge-Based Systems}, 256:109762, 2022.

\bibitem{gupta2022cyber-multi}
Lav Gupta, Tara Salman, Ali Ghubaish, Devrim Unal, Abdulla~Khalid Al-Ali, and Raj Jain.
\newblock Cybersecurity of multi-cloud healthcare systems: A hierarchical deep learning approach.
\newblock {\em Applied Soft Computing}, 118:108439, 2022.

\bibitem{gaddam2019anomaly}
Anuroop Gaddam, Tim Wilkin, and Maia Angelova.
\newblock Anomaly detection models for detecting sensor faults and outliers in the iot-a survey.
\newblock In {\em 2019 13th International Conference on Sensing Technology (ICST)}, pages 1--6. IEEE, 2019.

\bibitem{kamat2020anomaly}
Pooja Kamat and Rekha Sugandhi.
\newblock Anomaly detection for predictive maintenance in industry 4.0-a survey.
\newblock In {\em E3S web of conferences}, volume 170, page 02007. EDP Sciences, 2020.

\bibitem{schwendemann2021survey}
Sebastian Schwendemann, Zubair Amjad, and Axel Sikora.
\newblock A survey of machine-learning techniques for condition monitoring and predictive maintenance of bearings in grinding machines.
\newblock {\em Computers in Industry}, 125:103380, 2021.

\bibitem{kang2020machine}
Ziqiu Kang, Cagatay Catal, and Bedir Tekinerdogan.
\newblock Machine learning applications in production lines: A systematic literature review.
\newblock {\em Computers \& Industrial Engineering}, 149:106773, 2020.

\bibitem{davari2021survey}
Narjes Davari, Bruno Veloso, Gustavo de~Assis Costa, Pedro~Mota Pereira, Rita~P Ribeiro, and Jo{\~a}o Gama.
\newblock A survey on data-driven predictive maintenance for the railway industry.
\newblock {\em Sensors}, 21(17):5739, 2021.

\bibitem{nor2021overview}
Ahmad Kamal~Mohd Nor, Srinivasa~Rao Pedapati, Masdi Muhammad, and V{\'\i}ctor Leiva.
\newblock Overview of explainable artificial intelligence for prognostic and health management of industrial assets based on preferred reporting items for systematic reviews and meta-analyses.
\newblock {\em Sensors}, 21(23):8020, 2021.

\bibitem{petersen2008systematic}
Kai Petersen, Robert Feldt, Shahid Mujtaba, and Michael Mattsson.
\newblock Systematic mapping studies in software engineering.
\newblock In {\em 12th International Conference on Evaluation and Assessment in Software Engineering (EASE) 12}, pages 1--10, 2008.

\bibitem{mendonca2019disaster}
Julio Mendonca, Ermeson Andrade, Patricia~Takako Endo, and Ricardo Lima.
\newblock Disaster recovery solutions for it systems: A systematic mapping study.
\newblock {\em Journal of Systems and Software}, 149:511--530, 2019.

\bibitem{kitchenham2007cross}
Barbara~A Kitchenham, Emilia Mendes, and Guilherme~H Travassos.
\newblock Cross versus within-company cost estimation studies: A systematic review.
\newblock {\em IEEE Transactions on Software Engineering}, 33(5):316--329, 2007.

\bibitem{jakubowski2021anomaly}
Jakub Jakubowski, Przemys{\l}aw Stanisz, Szymon Bobek, and Grzegorz~J Nalepa.
\newblock Anomaly detection in asset degradation process using variational autoencoder and explanations.
\newblock {\em Sensors}, 22(1):291, 2021.

\bibitem{hong2022intelligent}
Seong~Hyeon Hong, Tristan Kyzer, Jackson Cornelius, Feraidoon Zahiri, and Yi~Wang.
\newblock Intelligent anomaly detection of robot manipulator based on energy consumption auditing.
\newblock In {\em 2022 IEEE Aerospace Conference (AERO)}, pages 1--11. IEEE, 2022.

\bibitem{assafo2021topsis}
Maryam Assafo and Peter Langend{\"o}rfer.
\newblock A topsis-assisted feature selection scheme and som-based anomaly detection for milling tools under different operating conditions.
\newblock {\em IEEE Access}, 9:90011--90028, 2021.

\bibitem{zabinski2021fpga}
Tomasz {\.Z}abi{\'n}ski, Zbigniew Hajduk, Jacek Kluska, and Les{\l}aw Gniewek.
\newblock Fpga-embedded anomaly detection system for milling process.
\newblock {\em IEEE Access}, 9:124059--124069, 2021.

\bibitem{ryu2022quantile}
Seunghyoung Ryu, Jiyeon Yim, Junghoon Seo, Yonggyun Yu, and Hogeon Seo.
\newblock Quantile autoencoder with abnormality accumulation for anomaly detection of multivariate sensor data.
\newblock {\em IEEE Access}, 10:70428--70439, 2022.

\bibitem{guo2022unsupervised}
Liang Guo, Yaoxiang Yu, Andongzhe Duan, Hongli Gao, and Jiangquan Zhang.
\newblock An unsupervised feature learning based health indicator construction method for performance assessment of machines.
\newblock {\em Mechanical Systems and Signal Processing}, 167:108573, 2022.

\bibitem{zhai2021enabling}
Simon Zhai, Benedikt Gehring, and Gunther Reinhart.
\newblock Enabling predictive maintenance integrated production scheduling by operation-specific health prognostics with generative deep learning.
\newblock {\em Journal of Manufacturing Systems}, 61:830--855, 2021.

\bibitem{watanabe2020anomaly}
Tsubasa Watanabe, Ippei Kono, and Hideaki Onozuka.
\newblock Anomaly detection methods in turning based on motor data analysis.
\newblock {\em Procedia Manufacturing}, 48:882--893, 2020.

\bibitem{kim2022anomaly}
Doyun Kim and Tae-Young Heo.
\newblock Anomaly detection with feature extraction based on machine learning using hydraulic system iot sensor data.
\newblock {\em Sensors}, 22(7):2479, 2022.

\bibitem{abbasi2021outliernets}
Saad Abbasi, Mahmoud Famouri, Mohammad~Javad Shafiee, and Alexander Wong.
\newblock Outliernets: Highly compact deep autoencoder network architectures for on-device acoustic anomaly detection.
\newblock {\em Sensors}, 21(14):4805, 2021.

\bibitem{boons2021edge}
Bert Boons, Marian Verhelst, and Peter Karsmakers.
\newblock Low power on-line machine monitoring at the edge.
\newblock In {\em 2021 International Conference on Applied Artificial Intelligence (ICAPAI)}, pages 1--8, 2021.

\bibitem{di2022anomalous}
Emanuele Di~Fiore, Antonino Ferraro, Antonio Galli, Vincenzo Moscato, and Giancarlo Sperl{\`\i}.
\newblock An anomalous sound detection methodology for predictive maintenance.
\newblock {\em Expert Systems with Applications}, 209:118324, 2022.

\bibitem{tagawa2021acoustic}
Yuki Tagawa, Rytis Maskeli{\=u}nas, and Robertas Dama{\v{s}}evi{\v{c}}ius.
\newblock Acoustic anomaly detection of mechanical failures in noisy real-life factory environments.
\newblock {\em Electronics}, 10(19):2329, 2021.

\bibitem{juodelyte2022predicting}
Dovile Juodelyte, Veronika Cheplygina, Therese Graversen, and Philippe Bonnet.
\newblock Predicting bearings' degradation stages for predictive maintenance in the pharmaceutical industry.
\newblock {\em arXiv preprint arXiv:2203.03259}, 2022.

\bibitem{song2019anomaly}
Na~Song, Xiangzhi Hu, and Ning Li.
\newblock Anomaly detection of wind turbine generator based on temporal information.
\newblock In {\em Proceedings of the 2019 7th International Conference on Information Technology: IoT and Smart City}, pages 477--482, 2019.

\bibitem{konig2021machine}
F~K{\"o}nig, C~Sous, A~Ouald Chaib, and G~Jacobs.
\newblock Machine learning based anomaly detection and classification of acoustic emission events for wear monitoring in sliding bearing systems.
\newblock {\em Tribology International}, 155:106811, 2021.

\bibitem{saxena2008damage}
Abhinav Saxena, Kai Goebel, and Donald~L Simon.
\newblock Damage propagation modeling for aircraft engine run-to-failure simulation.
\newblock {\em Aerospace Science and Technology}, 12(5):399--408, 2008.

\bibitem{bose2019adepos}
Sumon~Kumar Bose, Bapi Kar, Mohendra Roy, Pradeep~Kumar Gopalakrishnan, and Arindam Basu.
\newblock Adepos: Anomaly detection based power saving for predictive maintenance using edge computing.
\newblock In {\em Proceedings of the 24th asia and south pacific design automation conference}, pages 597--602, 2019.

\bibitem{nguyen2023time}
Thi Phuong~Quyen Nguyen, Phan Nguyen~Ky Phuc, Chao-Lung Yang, Hendri Sutrisno, Bao-Han Luong, Thi Huynh~Anh Le, and Thanh~Tung Nguyen.
\newblock Time-series anomaly detection using dynamic programming based longest common subsequence on sensor data.
\newblock {\em Expert Systems with Applications}, 213:118902, 2023.

\bibitem{patra2022anomaly}
Krishna Patra, Rabi~Narayan Sethi, and Dhiren~Kkumar Behera.
\newblock Anomaly detection in rotating machinery using autoencoders based onbidirectional lstm and gru neural networks.
\newblock {\em Turkish Journal of Electrical Engineering and Computer Sciences}, 30(4):1637--1653, 2022.

\bibitem{mostafavi2021novel}
Alireza Mostafavi and Ali Sadighi.
\newblock A novel online machine learning approach for real-time condition monitoring of rotating machines.
\newblock In {\em 2021 9th RSI International Conference on Robotics and Mechatronics (ICRoM)}, pages 267--273, 2021.

\bibitem{jiang2022multiscale}
Guoqian Jiang, Shiqiang Nie, Ping Xie, Yingwei Li, and Xiaoli Li.
\newblock Multiscale one-class classification network for machine health monitoring.
\newblock {\em IEEE Sensors Journal}, 22(13):13043--13054, 2022.

\bibitem{yu2022fastadaptation}
Yi-Cheng Yu, Shang-Wen Chuang, Hong-Han Shuai, and Chen-Yi Lee.
\newblock Fast adaption for multi motor anomaly detection via meta learning and deep unsupervised learning.
\newblock In {\em 2022 IEEE 31st International Symposium on Industrial Electronics (ISIE)}, pages 1186--1189, 2022.

\bibitem{kilian2022vibration}
Kilian Vos, Zhongxiao Peng, Christopher Jenkins, Md~Rifat Shahriar, Pietro Borghesani, and Wenyi Wang.
\newblock Vibration-based anomaly detection using lstm/svm approaches.
\newblock {\em Mechanical Systems and Signal Processing}, 169:108752, 2022.

\bibitem{brito2022explainable}
Lucas~C Brito, Gian~Antonio Susto, Jorge~N Brito, and Marcus~AV Duarte.
\newblock An explainable artificial intelligence approach for unsupervised fault detection and diagnosis in rotating machinery.
\newblock {\em Mechanical Systems and Signal Processing}, 163:108105, 2022.

\bibitem{jian2021initial}
Sirui Jian, Shigemi Ishida, and Yutaka Arakawa.
\newblock Initial attempt on wi-fi csi based vibration sensing for factory equipment fault detection.
\newblock In {\em Adjunct Proceedings of the 2021 International Conference on Distributed Computing and Networking}, pages 163--168, 2021.

\bibitem{de2022hybrid}
Rodrigo de~Paula~Monteiro, Mariela~Cerrada Lozada, Diego Roman~Cabrera Mendieta, Ren{\'e} Vinicio~S{\'a}nchez Loja, and Carmelo Jos{\'e}~Albanez Bastos~Filho.
\newblock A hybrid prototype selection-based deep learning approach for anomaly detection in industrial machines.
\newblock {\em Expert Systems with Applications}, page 117528, 2022.

\bibitem{li2020lifelong}
Chenyang Li, Lingfei Mo, Hanru Tang, and Ruqiang Yan.
\newblock Lifelong condition monitoring based on nb-iot for anomaly detection of machinery equipment.
\newblock {\em Procedia Manufacturing}, 49:144--149, 2020.

\bibitem{pittino2020automatic}
Federico Pittino, Michael Puggl, Thomas Moldaschl, and Christina Hirschl.
\newblock Automatic anomaly detection on in-production manufacturing machines using statistical learning methods.
\newblock {\em Sensors}, 20(8):2344, 2020.

\bibitem{alfeo2020using}
Antonio~L Alfeo, Mario~GCA Cimino, Giuseppe Manco, Ettore Ritacco, and Gigliola Vaglini.
\newblock Using an autoencoder in the design of an anomaly detector for smart manufacturing.
\newblock {\em Pattern Recognition Letters}, 136:272--278, 2020.

\bibitem{yun2020development}
Huitaek Yun, Hanjun Kim, Eunseob Kim, and Martin~BG Jun.
\newblock Development of internal sound sensor using stethoscope and its applications for machine monitoring.
\newblock {\em Procedia Manufacturing}, 48:1072--1078, 2020.

\bibitem{ying2021hybrid}
Zhenzhi Ying, Liming Shu, Toru Kizaki, Masatoshi Iwama, and Naohiko Sugita.
\newblock Hybrid approach for onsite monitoring and anomaly detection of cutting tool life.
\newblock {\em Procedia CIRP}, 104:1541--1546, 2021.

\bibitem{meire2019comparison}
Maarten Meire and Peter Karsmakers.
\newblock Comparison of deep autoencoder architectures for real-time acoustic based anomaly detection in assets.
\newblock In {\em 2019 10th IEEE International Conference on Intelligent Data Acquisition and Advanced Computing Systems: Technology and Applications (IDAACS)}, volume~2, pages 786--790. IEEE, 2019.

\bibitem{ahn2021deep}
Hyojung Ahn and Inchoon Yeo.
\newblock Deep-learning-based approach to anomaly detection techniques for large acoustic data in machine operation.
\newblock {\em Sensors}, 21(16):5446, 2021.

\bibitem{gruner2020evaluation}
Tobias Gr{\"u}ner, Falco B{\"o}llhoff, Robert Meisetschl{\"a}ger, Alexander Vydrenko, Martyna Bator, Alexander Dicks, and Andreas Theissler.
\newblock Evaluation of machine learning for sensorless detection and classification of faults in electromechanical drive systems.
\newblock {\em Procedia Computer Science}, 176:1586--1595, 2020.

\bibitem{hiruta2021unsupervised}
Tomoaki Hiruta, Kohji Maki, Tetsuji Kato, and Yasushi Umeda.
\newblock Unsupervised learning based diagnosis model for anomaly detection of motor bearing with current data.
\newblock {\em Procedia CIRP}, 98:336--341, 2021.

\bibitem{givnan2022anomaly}
Sean Givnan, Carl Chalmers, Paul Fergus, Sandra Ortega-Martorell, and Tom Whalley.
\newblock Anomaly detection using autoencoder reconstruction upon industrial motors.
\newblock {\em Sensors}, 22(9):3166, 2022.

\bibitem{park2021design}
YeongHyeon Park and Myung~Jin Kim.
\newblock Design of cost-effective auto-encoder for electric motor anomaly detection in resource constrained edge device.
\newblock In {\em 2021 IEEE 3rd Eurasia Conference on IOT, Communication and Engineering (ECICE)}, pages 241--246. IEEE, 2021.

\bibitem{panagou2022feature}
Sotirios Panagou, Fabio Fruggiero, Marida Lerra, Carmen del Vecchio, Fernando Menchetti, Luca Piedimonte, Oreste~Riccardo Natale, and Salvatore Passariello.
\newblock Feature investigation with digital twin for predictive maintenance following a machine learning approach.
\newblock {\em IFAC-PapersOnLine}, 55(2):132--137, 2022.

\bibitem{yanabe2020anomaly}
Tomu Yanabe, Hiroaki Nishi, and Masahiro Hashimoto.
\newblock Anomaly detection based on histogram methodology and factor analysis using lightgbm for cooling systems.
\newblock In {\em 2020 25th IEEE International Conference on Emerging Technologies and Factory Automation (ETFA)}, volume~1, pages 952--958. IEEE, 2020.

\bibitem{gonzalez2022two}
Ana Gonz{\'a}lez-Mu{\~n}iz, Ignacio D{\'\i}az, Abel~A Cuadrado, Diego Garc{\'\i}a-P{\'e}rez, and Daniel P{\'e}rez.
\newblock Two-step residual-error based approach for anomaly detection in engineering systems using variational autoencoders.
\newblock {\em Computers and Electrical Engineering}, 101:108065, 2022.

\bibitem{velasquez2022hybrid}
David Velasquez, Enrique Perez, Xabier Oregui, Arkaitz Artetxe, Jorge Manteca, Jordi~Escayola Mansilla, Mauricio Toro, Mikel Maiza, and Basilio Sierra.
\newblock A hybrid machine-learning ensemble for anomaly detection in real-time industry 4.0 systems.
\newblock {\em IEEE Access}, 10:72024--72036, 2022.

\bibitem{canizo2019multi}
Mikel Canizo, Isaac Triguero, Angel Conde, and Enrique Onieva.
\newblock Multi-head cnn--rnn for multi-time series anomaly detection: An industrial case study.
\newblock {\em Neurocomputing}, 363:246--260, 2019.

\bibitem{conradi2021anomaly}
Jos{\'e}~Luis Conradi~Hoffmann, Leonardo~Passig Horstmann, Mateus Mart{\'\i}nez~Lucena, Gustavo Medeiros~de Araujo, Ant{\^o}nio~Augusto Fr{\"o}hlich, and Marcos Hisashi~Napoli Nishioka.
\newblock Anomaly detection on wind turbines based on a deep learning analysis of vibration signals.
\newblock {\em Applied Artificial Intelligence}, 35(12):893--913, 2021.

\bibitem{yan2023hybrid}
Shen Yan, Haidong Shao, Yiming Xiao, Bin Liu, and Jiafu Wan.
\newblock Hybrid robust convolutional autoencoder for unsupervised anomaly detection of machine tools under noises.
\newblock {\em Robotics and Computer-Integrated Manufacturing}, 79:102441, 2023.

\bibitem{denkena2021data}
B~Denkena, M-A Dittrich, H~Noske, D~Stoppel, and D~Lange.
\newblock Data-based ensemble approach for semi-supervised anomaly detection in machine tool condition monitoring.
\newblock {\em CIRP Journal of Manufacturing Science and Technology}, 35:795--802, 2021.

\bibitem{yang2022interpretable}
Wei-Ting Yang, Marco~S Reis, Valeria Borodin, Michel Juge, and Agn{\`e}s Roussy.
\newblock An interpretable unsupervised bayesian network model for fault detection and diagnosis.
\newblock {\em Control Engineering Practice}, 127:105304, 2022.

\bibitem{apostol2021change}
Elena-Simona Apostol, Ciprian-Octavian Truic{\u{a}}, Florin Pop, and Christian Esposito.
\newblock Change point enhanced anomaly detection for iot time series data.
\newblock {\em Water}, 13(12):1633, 2021.

\bibitem{lindemann2019anomaly}
Benjamin Lindemann, Fabian Fesenmayr, Nasser Jazdi, and Michael Weyrich.
\newblock Anomaly detection in discrete manufacturing using self-learning approaches.
\newblock {\em Procedia CIRP}, 79:313--318, 2019.

\bibitem{meyer2022anomaly}
Kevin Meyer and Vladimir Mahalec.
\newblock Anomaly detection methods for infrequent failures in resistive steel welding.
\newblock {\em Journal of Manufacturing Processes}, 75:497--513, 2022.

\bibitem{kammerer2019anomaly}
Klaus Kammerer, Burkhard Hoppenstedt, R{\"u}diger Pryss, Steffen St{\"o}kler, Johannes Allgaier, and Manfred Reichert.
\newblock Anomaly detections for manufacturing systems based on sensor data—insights into two challenging real-world production settings.
\newblock {\em Sensors}, 19(24):5370, 2019.

\bibitem{langone2020interpretable}
Rocco Langone, Alfredo Cuzzocrea, and Nikolaos Skantzos.
\newblock Interpretable anomaly prediction: Predicting anomalous behavior in industry 4.0 settings via regularized logistic regression tools.
\newblock {\em Data \& Knowledge Engineering}, 130:101850, 2020.

\bibitem{calvo2023collaborative}
Pablo Calvo-Bascones, Alexandre Voisin, Phuc Do, and Miguel~A Sanz-Bobi.
\newblock A collaborative network of digital twins for anomaly detection applications of complex systems. snitch digital twin concept.
\newblock {\em Computers in Industry}, 144:103767, 2023.

\bibitem{abbracciavento2021anomaly}
Francesco Abbracciavento, Simone Formentin, Jacopo Balocco, Andrea Rota, Vincenzo Manzoni, and Sergio~M Savaresi.
\newblock Anomaly detection via distributed sensing: a var modeling approach.
\newblock {\em IFAC-PapersOnLine}, 54(7):85--90, 2021.

\bibitem{wang2016self}
Xing Wang, Jessica Lin, Nital Patel, and Martin Braun.
\newblock A self-learning and online algorithm for time series anomaly detection, with application in cpu manufacturing.
\newblock In {\em Proceedings of the 25th ACM International on Conference on Information and Knowledge Management}, pages 1823--1832, 2016.

\bibitem{wielgosz2019mapping}
Maciej Wielgosz and Micha{\l} Karwatowski.
\newblock Mapping neural networks to fpga-based iot devices for ultra-low latency processing.
\newblock {\em Sensors}, 19(13):2981, 2019.

\bibitem{keleko2023health}
Aurelien~Teguede Keleko, Bernard Kamsu-Foguem, Raymond~Houe Ngouna, and Am{\`e}vi Tongne.
\newblock Health condition monitoring of a complex hydraulic system using deep neural network and deepshap explainable xai.
\newblock {\em Advances in Engineering Software}, 175:103339, 2023.

\bibitem{coelho2022predictive}
Daniel Coelho, Diogo Costa, Eug{\'e}nio~M Rocha, Duarte Almeida, and Jos{\'e}~P Santos.
\newblock Predictive maintenance on sensorized stamping presses by time series segmentation, anomaly detection, and classification algorithms.
\newblock {\em Procedia Computer Science}, 200:1184--1193, 2022.

\bibitem{tancredi2022integration}
Giovanni~Paolo Tancredi, Giuseppe Vignali, and Eleonora Bottani.
\newblock Integration of digital twin, machine-learning and industry 4.0 tools for anomaly detection: An application to a food plant.
\newblock {\em Sensors}, 22(11):4143, 2022.

\bibitem{de2022long}
Anthony~Fombonne de~Galatheau, Alexandru-Liviu Olteanu, Nathalie Julien, and Steven Le~Garrec.
\newblock Long short term memory-based anomaly detection applied to an industrial dosing pump.
\newblock {\em IFAC-PapersOnLine}, 55(2):240--245, 2022.

\bibitem{liu2021online}
Keying Liu, Wentao Mao, Huadong Shi, Chao Wu, and Jiaxian Chen.
\newblock Online anomaly detection with streaming data based on fine-grained feature forecasting.
\newblock In {\em 2021 33rd Chinese Control and Decision Conference (CCDC)}, pages 454--459, 2021.

\bibitem{antonini2022tinyml}
Mattia Antonini, Miguel Pincheira, Massimo Vecchio, and Fabio Antonelli.
\newblock A tinyml approach to non-repudiable anomaly detection in extreme industrial environments.
\newblock In {\em 2022 IEEE International Workshop on Metrology for Industry 4.0 \& IoT (MetroInd4. 0\&IoT)}, pages 397--402. IEEE, 2022.

\bibitem{zufle2022workflow}
Marwin Züfle, Felix Moog, Veronika Lesch, Christian Krupitzer, and Samuel Kounev.
\newblock A machine learning-based workflow for automatic detection of anomalies in machine tools.
\newblock {\em ISA Transactions}, 125:445--458, 2022.

\bibitem{yasaei2020iot}
Rozhin Yasaei, Felix Hernandez, and Mohammad Abdullah~Al Faruque.
\newblock Iot-cad: Context-aware adaptive anomaly detection in iot systems through sensor association.
\newblock In {\em Proceedings of the 39th International Conference on Computer-Aided Design}, pages 1--9, 2020.

\bibitem{de2020use}
Fabrizio De~Vita, Dario Bruneo, and Sajal~K Das.
\newblock On the use of a full stack hardware/software infrastructure for sensor data fusion and fault prediction in industry 4.0.
\newblock {\em Pattern Recognition Letters}, 138:30--37, 2020.

\bibitem{hendrickx2020general}
Kilian Hendrickx, Wannes Meert, Yves Mollet, Johan Gyselinck, Bram Cornelis, Konstantinos Gryllias, and Jesse Davis.
\newblock A general anomaly detection framework for fleet-based condition monitoring of machines.
\newblock {\em Mechanical Systems and Signal Processing}, 139:106585, 2020.

\bibitem{kim2019online}
Chul Kim, Inwhee Joe, Deokwon Jang, Eunji Kim, and Sanghun Nam.
\newblock Online monitoring automation using anomaly detection in iot/it environment.
\newblock In {\em Computer Science On-line Conference}, pages 96--106. Springer, 2019.

\bibitem{calvo2021anomaly}
Pablo Calvo-Bascones, Miguel~A Sanz-Bobi, and Thomas~M Welte.
\newblock Anomaly detection method based on the deep knowledge behind behavior patterns in industrial components. application to a hydropower plant.
\newblock {\em Computers in Industry}, 125:103376, 2021.

\bibitem{lu2021gan}
Haodong Lu, Miao Du, Kai Qian, Xiaoming He, and Kun Wang.
\newblock Gan-based data augmentation strategy for sensor anomaly detection in industrial robots.
\newblock {\em IEEE Sensors Journal}, 22(18):17464--17474, 2021.

\bibitem{shi2021deep}
Huadong Shi, Wentao Mao, Gangsheng Wang, and Keying Liu.
\newblock Deep multi-task svdd: A new robust online detection method of bearings early fault.
\newblock In {\em 2021 Global Reliability and Prognostics and Health Management (PHM-Nanjing)}, pages 1--7. IEEE, 2021.

\bibitem{ds2022comparative}
Bhupal~Naik DS, Venkatesulu Dondeti, and Sivadi Balakrishna.
\newblock Comparative analysis of machine learning-based algorithms for detection of anomalies in iiot.
\newblock {\em International Journal of Information Retrieval Research (IJIRR)}, 12(1):1--55, 2022.

\bibitem{luo2022multi}
Qinyuan Luo, Jinglong Chen, Yanyang Zi, Yuanhong Chang, and Yong Feng.
\newblock Multi-mode non-gaussian variational autoencoder network with missing sources for anomaly detection of complex electromechanical equipment.
\newblock {\em ISA transactions}, 2022.

\bibitem{choi2022explainable}
Heejeong Choi, Donghwa Kim, Jounghee Kim, Jina Kim, and Pilsung Kang.
\newblock Explainable anomaly detection framework for predictive maintenance in manufacturing systems.
\newblock {\em Applied Soft Computing}, 125:109147, 2022.

\bibitem{qian2021multichannel}
Qingting Qian, Xiaolei Fang, Jinwu Xu, and Min Li.
\newblock Multichannel profile-based monitoring method and its application in the basic oxygen furnace steelmaking process.
\newblock {\em Journal of Manufacturing Systems}, 61:375--390, 2021.

\bibitem{iftikhar2020outlier}
Nadeem Iftikhar, Thorkil Baattrup-Andersen, Finn~Ebertsen Nordbjerg, and Karsten Jeppesen.
\newblock Outlier detection in sensor data using ensemble learning.
\newblock {\em Procedia Computer Science}, 176:1160--1169, 2020.

\bibitem{wang2021early}
Baicun Wang, Yang Li, Ying Luo, Xingyu Li, and Theodor Freiheit.
\newblock Early event detection in a deep-learning driven quality prediction model for ultrasonic welding.
\newblock {\em Journal of Manufacturing Systems}, 60:325--336, 2021.

\bibitem{wu2019lstm}
Di~Wu, Zhongkai Jiang, Xiaofeng Xie, Xuetao Wei, Weiren Yu, and Renfa Li.
\newblock Lstm learning with bayesian and gaussian processing for anomaly detection in industrial iot.
\newblock {\em IEEE Transactions on Industrial Informatics}, 16(8):5244--5253, 2019.

\bibitem{wang2022detecting}
Yue Wang, Michael Perry, Dane Whitlock, and John~W Sutherland.
\newblock Detecting anomalies in time series data from a manufacturing system using recurrent neural networks.
\newblock {\em Journal of Manufacturing Systems}, 62:823--834, 2022.

\bibitem{netzer2022process}
Markus Netzer, Jannik Bach, Alexander Puchta, Philipp G{\"o}nnheimer, and J{\"u}rgen Fleischer.
\newblock Process segmented based intelligent anomaly detection in highly flexible production machines under low machine data availability.
\newblock {\em Procedia CIRP}, 107:647--652, 2022.

\bibitem{lughofer2020line}
Edwin Lughofer, Alexandru-Ciprian Zavoianu, Robert Pollak, Mahardhika Pratama, Pauline Meyer-Heye, Helmut Z{\"o}rrer, Christian Eitzinger, and Thomas Radauer.
\newblock On-line anomaly detection with advanced independent component analysis of multi-variate residual signals from causal relation networks.
\newblock {\em Information Sciences}, 537:425--451, 2020.

\bibitem{li2022correlation}
Han Li, Xinyu Wang, Zhongguo Yang, Sikandar Ali, Ning Tong, and Samad Baseer.
\newblock Correlation-based anomaly detection method for multi-sensor system.
\newblock {\em Computational Intelligence and Neuroscience}, 2022, 2022.

\bibitem{kabadayi2006virtual}
Sanem Kabadayi, Adam Pridgen, and Christine Julien.
\newblock Virtual sensors: Abstracting data from physical sensors.
\newblock In {\em 2006 International Symposium on a World of Wireless, Mobile and Multimedia Networks (WoWMoM'06)}, pages 6--pp. IEEE, 2006.

\bibitem{lai2020full}
Zhilu Lai, Ignacio Alzugaray, Margarita Chli, and Eleni Chatzi.
\newblock Full-field structural monitoring using event cameras and physics-informed sparse identification.
\newblock {\em Mechanical Systems and Signal Processing}, 145:106905, 2020.

\bibitem{geron2022hands}
Aur{\'e}lien G{\'e}ron.
\newblock {\em Hands-on machine learning with Scikit-Learn, Keras, and TensorFlow}.
\newblock " O'Reilly Media, Inc.", 2022.

\bibitem{baldi2012autoencoders}
Pierre Baldi.
\newblock Autoencoders, unsupervised learning, and deep architectures.
\newblock In {\em Proceedings of ICML workshop on unsupervised and transfer learning}, pages 37--49. JMLR Workshop and Conference Proceedings, 2012.

\bibitem{wickramasinghe2021resnet}
Chathurika~S Wickramasinghe, Daniel~L Marino, and Milos Manic.
\newblock Resnet autoencoders for unsupervised feature learning from high-dimensional data: Deep models resistant to performance degradation.
\newblock {\em IEEE Access}, 9:40511--40520, 2021.

\bibitem{ali2016basic}
Zulfiqar Ali and S~Bala Bhaskar.
\newblock Basic statistical tools in research and data analysis.
\newblock {\em Indian journal of anaesthesia}, 60(9):662, 2016.

\bibitem{aggarwal2011noise}
Rajeev Aggarwal, Jai~Karan Singh, Vijay~Kumar Gupta, Sanjay Rathore, Mukesh Tiwari, and Anubhuti Khare.
\newblock Noise reduction of speech signal using wavelet transform with modified universal threshold.
\newblock {\em International Journal of Computer Applications}, 20(5):14--19, 2011.

\bibitem{groth2013principal}
Detlef Groth, Stefanie Hartmann, Sebastian Klie, and Joachim Selbig.
\newblock Principal components analysis.
\newblock {\em Computational Toxicology: Volume II}, pages 527--547, 2013.

\bibitem{smolinska2014current}
Agnieszka Smolinska, A-Ch Hauschild, RRR Fijten, JW~Dallinga, Jan Baumbach, and FJ~Van~Schooten.
\newblock Current breathomics—a review on data pre-processing techniques and machine learning in metabolomics breath analysis.
\newblock {\em Journal of breath research}, 8(2):027105, 2014.

\bibitem{jogin2018feature}
Manjunath Jogin, MS~Madhulika, GD~Divya, RK~Meghana, S~Apoorva, et~al.
\newblock Feature extraction using convolution neural networks (cnn) and deep learning.
\newblock In {\em 2018 3rd IEEE international conference on recent trends in electronics, information \& communication technology (RTEICT)}, pages 2319--2323. IEEE, 2018.

\bibitem{zhou2022contrastive}
Hao Zhou, Ke~Yu, Xuan Zhang, Guanlin Wu, and Anis Yazidi.
\newblock Contrastive autoencoder for anomaly detection in multivariate time series.
\newblock {\em Information Sciences}, 610:266--280, 2022.

\bibitem{liang2021robust}
Haoran Liang, Lei Song, Jianxing Wang, Lili Guo, Xuzhi Li, and Ji~Liang.
\newblock Robust unsupervised anomaly detection via multi-time scale dcgans with forgetting mechanism for industrial multivariate time series.
\newblock {\em Neurocomputing}, 423:444--462, 2021.

\bibitem{creswell2018generative}
Antonia Creswell, Tom White, Vincent Dumoulin, Kai Arulkumaran, Biswa Sengupta, and Anil~A Bharath.
\newblock Generative adversarial networks: An overview.
\newblock {\em IEEE signal processing magazine}, 35(1):53--65, 2018.

\bibitem{erfani2016high}
Sarah~M Erfani, Sutharshan Rajasegarar, Shanika Karunasekera, and Christopher Leckie.
\newblock High-dimensional and large-scale anomaly detection using a linear one-class svm with deep learning.
\newblock {\em Pattern Recognition}, 58:121--134, 2016.

\bibitem{liu2008isolation}
Fei~Tony Liu, Kai~Ming Ting, and Zhi-Hua Zhou.
\newblock Isolation forest.
\newblock In {\em 2008 eighth ieee international conference on data mining}, pages 413--422. IEEE, 2008.

\bibitem{cheng2019outlier}
Zhangyu Cheng, Chengming Zou, and Jianwei Dong.
\newblock Outlier detection using isolation forest and local outlier factor.
\newblock In {\em Proceedings of the conference on research in adaptive and convergent systems}, pages 161--168, 2019.

\bibitem{hossin2015review}
Mohammad Hossin and Md~Nasir Sulaiman.
\newblock A review on evaluation metrics for data classification evaluations.
\newblock {\em International journal of data mining \& knowledge management process}, 5(2):1, 2015.

\bibitem{zhao2022lithium}
Shaishai Zhao, Chaolong Zhang, and Yuanzhi Wang.
\newblock Lithium-ion battery capacity and remaining useful life prediction using board learning system and long short-term memory neural network.
\newblock {\em Journal of Energy Storage}, 52:104901, 2022.

\bibitem{berghout2022systematic}
Tarek Berghout and Mohamed Benbouzid.
\newblock A systematic guide for predicting remaining useful life with machine learning.
\newblock {\em Electronics}, 11(7), 2022.

\bibitem{siahpour2022novel}
Shahin Siahpour, Xiang Li, and Jay Lee.
\newblock A novel transfer learning approach in remaining useful life prediction for incomplete dataset.
\newblock {\em IEEE Transactions on Instrumentation and Measurement}, 71:1--11, 2022.

\bibitem{song2022novel}
Song Fu, Shisheng Zhong, Lin Lin, and Minghang Zhao.
\newblock A novel time-series memory auto-encoder with sequentially updated reconstructions for remaining useful life prediction.
\newblock {\em IEEE Transactions on Neural Networks and Learning Systems}, 33(12):7114--7125, 2022.

\bibitem{lin2020battery}
Chun-Pang Lin, Javier Cabrera, Fangfang Yang, Man-Ho Ling, Kwok-Leung Tsui, and Suk-Joo Bae.
\newblock Battery state of health modeling and remaining useful life prediction through time series model.
\newblock {\em Applied Energy}, 275:115338, 2020.

\bibitem{emmerich2018tutorial}
Michael~TM Emmerich and Andr{\'e}~H Deutz.
\newblock A tutorial on multiobjective optimization: fundamentals and evolutionary methods.
\newblock {\em Natural computing}, 17:585--609, 2018.

\bibitem{bauer2019real}
Felix~Christian Bauer, Dylan~Richard Muir, and Giacomo Indiveri.
\newblock Real-time ultra-low power ecg anomaly detection using an event-driven neuromorphic processor.
\newblock {\em IEEE transactions on biomedical circuits and systems}, 13(6):1575--1582, 2019.

\end{thebibliography}

\begin{IEEEbiography}[{\includegraphics[width=1in,height=1.25in,clip,keepaspectratio]{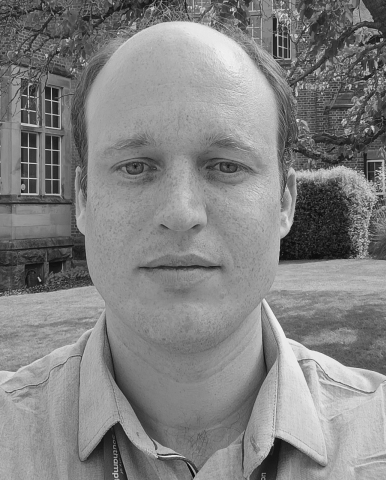}}]{S\'ergio F. Chevtchenko} received the M.Sc. degree in applied informatics from Universidade Federal Rural de Pernambuco (UFRPE) in 2018 and Ph.D. in computer science from Universidade Federal de Pernambuco (UFPE) in 2023. He is currently a researcher at the SENAI Institute of Innovation for Information and Communication Technologies (ISI-TICs). His research interests
include neuromorphic engineering, reinforcement learning and anomaly detection.
\end{IEEEbiography}

\begin{IEEEbiography}[{\includegraphics[width=1in,height=1.25in,clip,keepaspectratio]{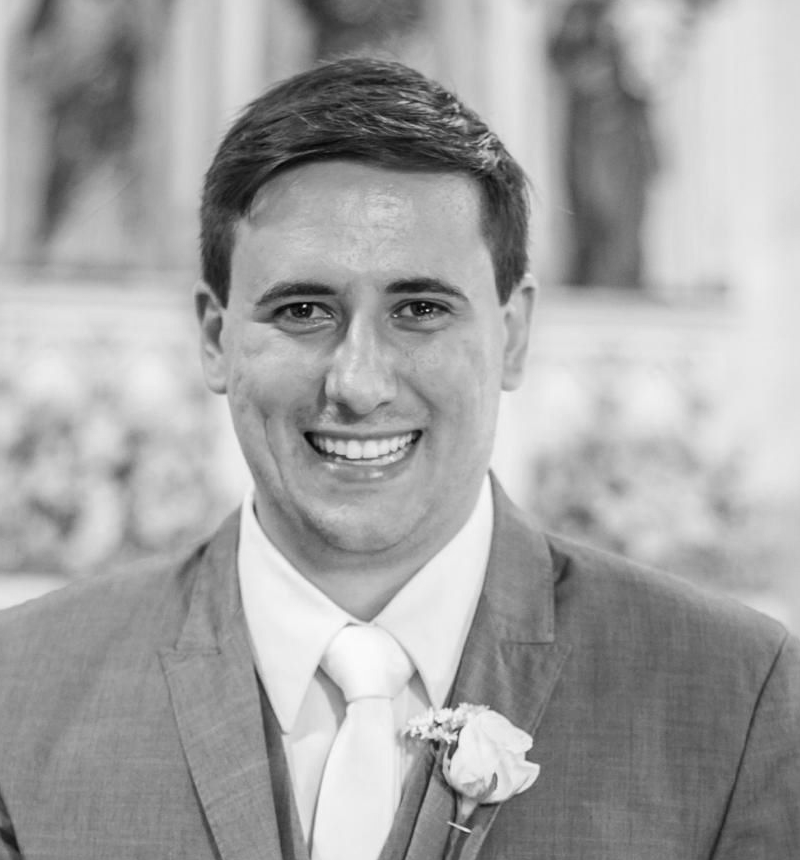}}]{Elisson da Silva Rocha} is a Ph.D. student in the Programa de Pós Graduação em Engenharia da Computação (PPGEC) at the Universidade de Pernambuco (UPE) and currently a researcher at the SENAI Institute of Innovation for Information and Communication Technologies (ISI-TICs), and the DotLAB Brazil Research Group. His areas of interest include Machine Learning and Computer Vision.
\end{IEEEbiography}

\begin{IEEEbiography}[{\includegraphics[width=1in,height=1.25in,clip,keepaspectratio]{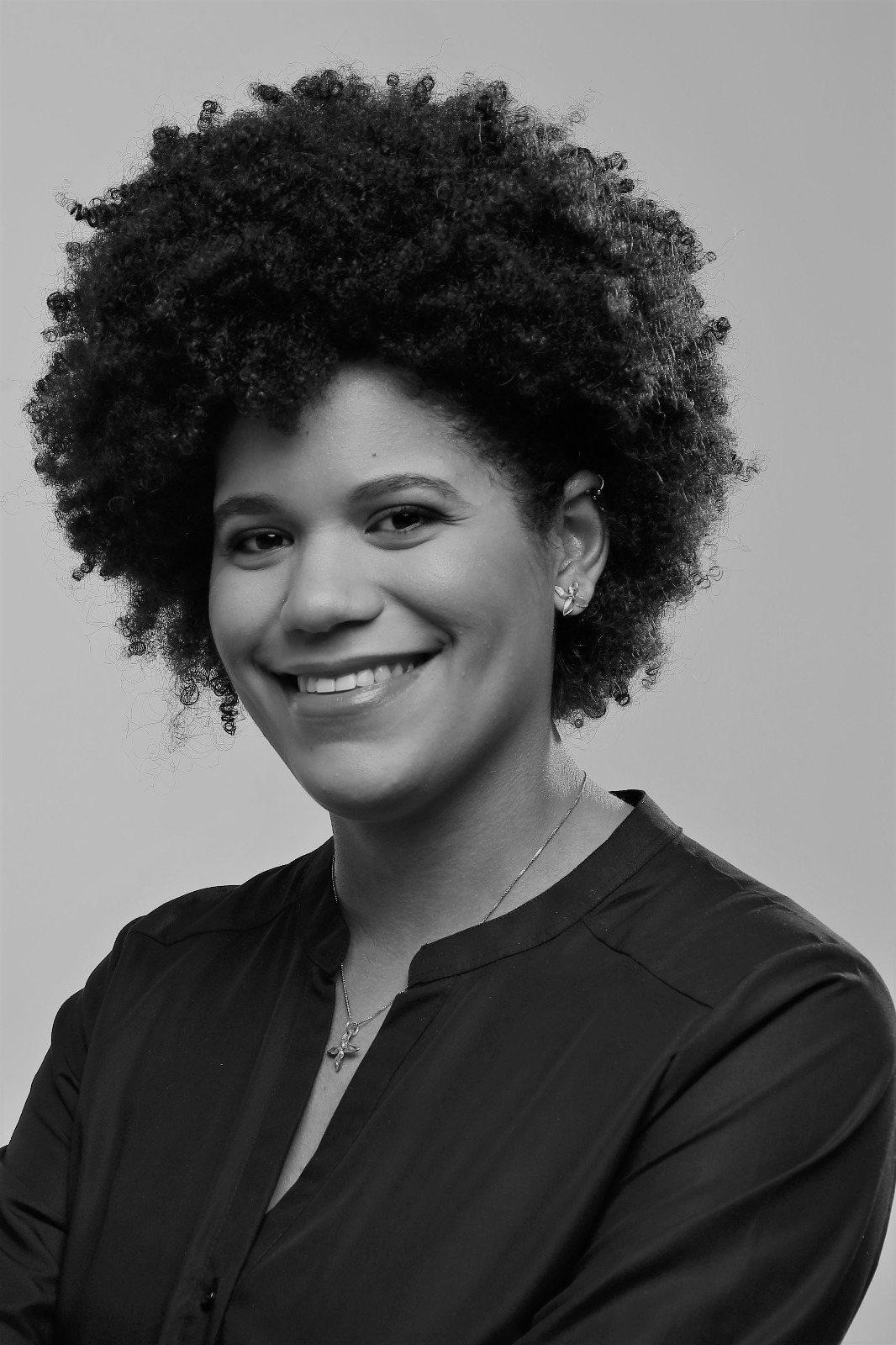}}]{Monalisa C. M. dos Santos}
is a researcher at SENAI Institute of Innovation for Information and Communication Technologies (ISI-TICs), Recife, Brazil, since 2021. Recieved a degree in Industrial Engineering from Federal University of Pernambuco (UFPE). Has a MsC dregree (2020) and is a PhD candidate in Industrial Engineering at UFPE. Her current research interest include  maintanence and reliability, artificial intelligence,and
computer vison.  
\end{IEEEbiography}

\begin{IEEEbiography}[{\includegraphics[width=1in,height=1.25in,clip,keepaspectratio]{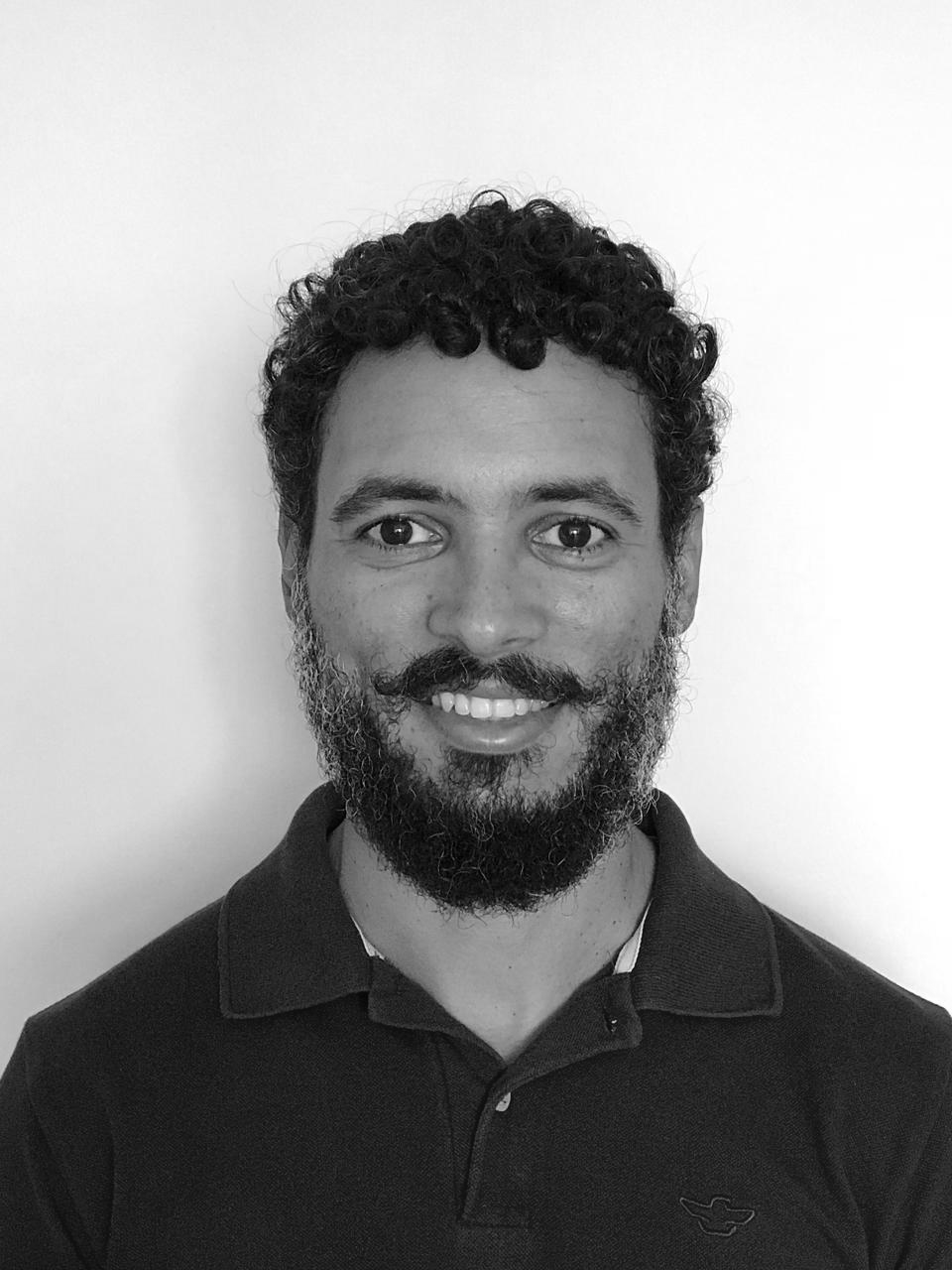}}]{Ricardo Lins Mota}
is a researcher at SENAI Institute of Innovation for Information and Communication Technologies (ISI-TICs). Received a M.Sc. degree in Computer Science from UFPE (Federal University of Pernambuco). His current research interest include  artificial intelligence, embedded systems and IoT.  
\end{IEEEbiography}

\begin{IEEEbiography}[{\includegraphics[width=1in,height=1.25in,clip,keepaspectratio]{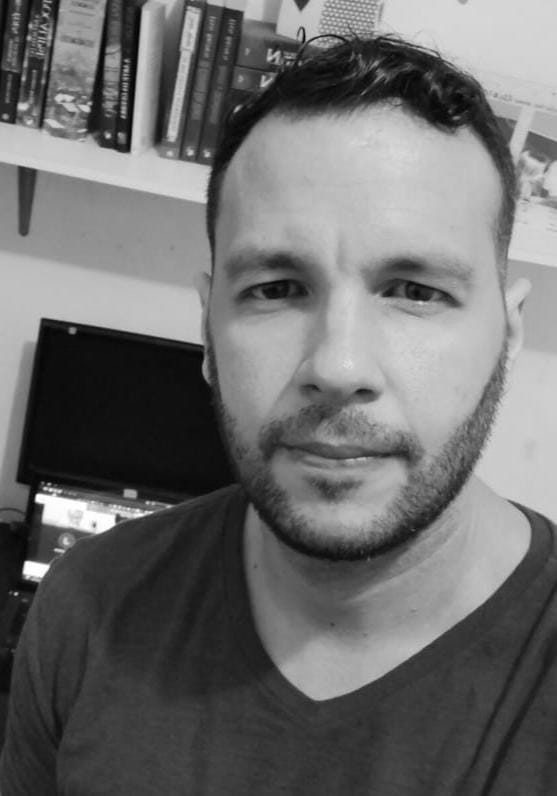}}]{Diego Moura Vieira}
received the B.S. in Telecommunication Engineering from Centro Universitário Maurício de Nassau, Recife, Pernambuco, in 2012. M.S. in Computer Science from Universidade Federal de Pernambuco in 2023. Since 2018, he has been Research Assistant at the SENAI Institute of Innovation for Information and Communication Technology (ISI-TICs). His research interests include IoT sensing networks and the application of AI in the Energy sector.  
\end{IEEEbiography}

\begin{IEEEbiography}[{\includegraphics[width=1in,height=1.25in,clip,keepaspectratio]{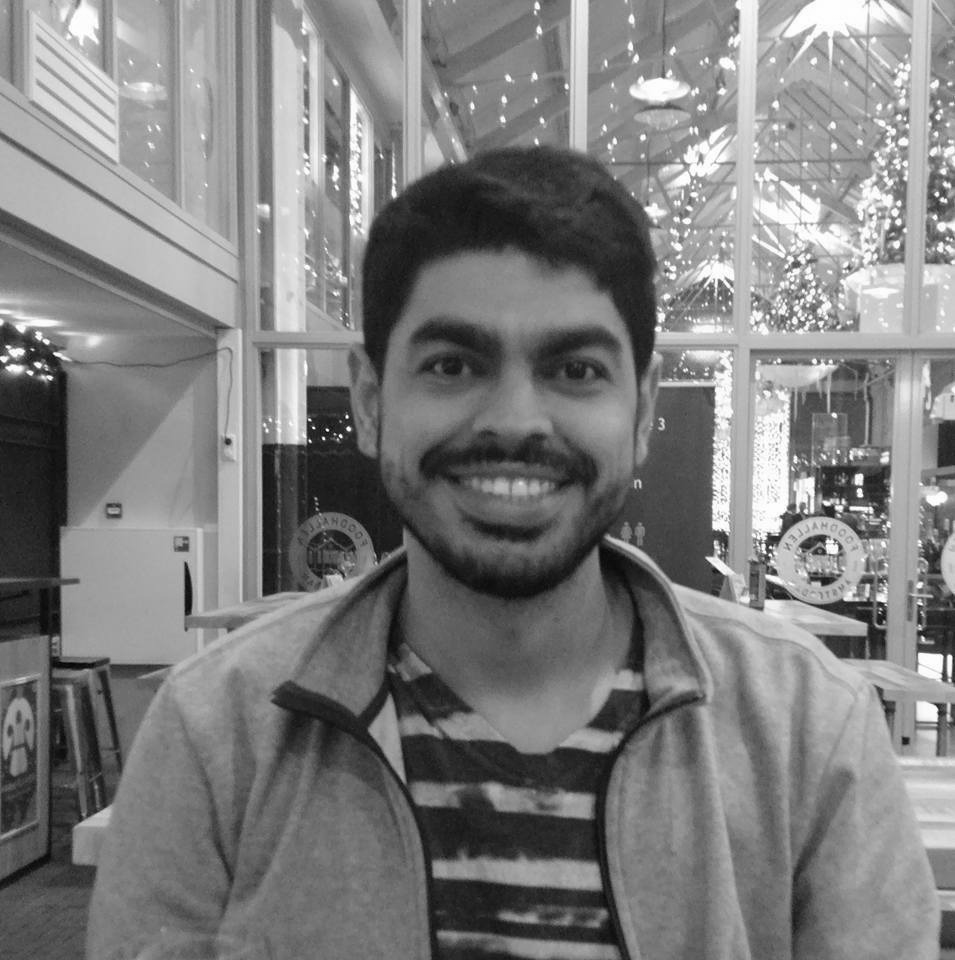}}]{Ermeson Carneiro de Andrade}
is an associate professor at the Department of Computing at the Federal Rural University of Pernambuco, Brazil. In 2014, he completed his PhD in Computer Science from the Federal University of Pernambuco. His research interest is focused in Performance, Dependability, and Modeling.  
\end{IEEEbiography}

\begin{IEEEbiography}[{\includegraphics[width=1in,height=1.25in,clip,keepaspectratio]{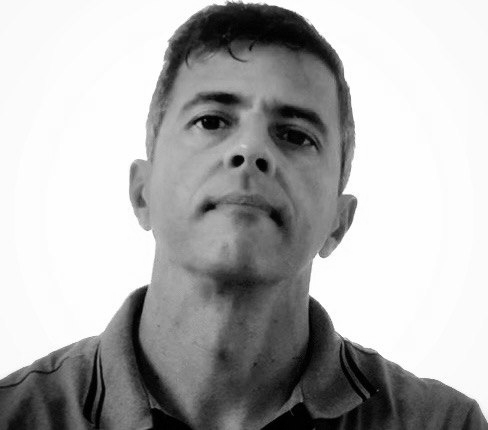}}]{Danilo Ricardo Barbosa de Araújo}
is an associate professor at the Department of Computing at the Federal Rural University of Pernambuco, Brazil. In 2015, he completed his PhD in Electrical Engineering from the Federal University of Pernambuco. His research interest is focused in Artificial Intelligence, Networking and Internet of Things.  
\end{IEEEbiography}

\EOD

\end{document}